\documentclass[10pt,journal,compsoc]{IEEEtran}
\usepackage{cite}
\ifCLASSINFOpdf
\else
\fi

\usepackage{xcolor}
\usepackage{booktabs}
\usepackage{epsfig}
\usepackage{graphicx}
\usepackage{amsmath}

\usepackage{amssymb}

\usepackage{verbatim}

\usepackage{url}
\usepackage{tabularx}
\usepackage{multirow}
\usepackage{subfigure}
\usepackage{amsmath,dsfont}
\usepackage{array}
\usepackage{enumitem}

\usepackage{pifont}
\usepackage{threeparttable}
\usepackage{bbding}

\usepackage{makecell}

\usepackage{verbatim}

\usepackage{colortbl}
\definecolor{mygray}{gray}{.75}

\usepackage[normalem]{ulem}

\usepackage{caption}

\begin{document}

%
% paper title
% Titles are generally capitalized except for words such as a, an, and, as,
% at, but, by, for, in, nor, of, on, or, the, to and up, which are usually
% not capitalized unless they are the first or last word of the title.
% Linebreaks \\ can be used within to get better formatting as desired.
% Do not put math or special symbols in the title.
% \title{A Survey on Recent Advances in Arabic Named Entity Recognition}
\title{A Survey on Arabic Named Entity Recognition:
Past, Recent Advances, and Future Trends}

%
%
% author names and IEEE memberships
% note positions of commas and nonbreaking spaces ( ~ ) LaTeX will not break
% a structure at a ~ so this keeps an author's name from being broken across
% two lines.
% use \thanks{} to gain access to the first footnote area
% a separate \thanks must be used for each paragraph as LaTeX2e's \thanks
% was not built to handle multiple paragraphs
%
%
%\IEEEcompsocitemizethanks is a special \thanks that produces the bulleted
% lists the Computer Society journals use for "first footnote" author
% affiliations. Use \IEEEcompsocthanksitem which works much like \item
% for each affiliation group. When not in compsoc mode,
% \IEEEcompsocitemizethanks becomes like \thanks and
% \IEEEcompsocthanksitem becomes a line break with idention. This
% facilitates dual compilation, although admittedly the differences in the
% desired content of \author between the different types of papers makes a
% one-size-fits-all approach a daunting prospect. For instance, compsoc 
% journal papers have the author affiliations above the "Manuscript
% received ..."  text while in non-compsoc journals this is reversed. Sigh.

% \author{Xiaoye Qu, XXX ~\IEEEmembership{Fellow,~IEEE}% <-this % stops a space
\author{Xiaoye Qu, Yingjie Gu, Qingrong Xia, Zechang Li, Zhefeng Wang, Baoxing Huai% <-this % stops a space
% \thanks{X. Qu,  are with Huawei Cloud, Hangzhou 310035, China. E-mail: quxiaoye@huawei.com}
\IEEEcompsocitemizethanks{\IEEEcompsocthanksitem X. Qu, Y.Gu, Q. Xia, Z. Li, Z. Wang, B. Huai are with Huawei Cloud System AI Innovation Lab, Hangzhou 310051, China. \protect E-mail: \{quxiaoye, guyingjie4, xiaqingrong, lizechang1, wangzhefeng, huaibaoxing\}@huawei.com\\
Zhefeng Wang is the corresponding author.}}
% \IEEEcompsocthanksitem xxx.}% <-this % stops an unwanted space
% \thanks{Manuscript received April 19, 2005; revised August 26, 2015.}}

% note the % following the last \IEEEmembership and also \thanks - 
% these prevent an unwanted space from occurring between the last author name
% and the end of the author line. i.e., if you had this:
% 
% \author{....lastname \thanks{...} \thanks{...} }
%                     ^------------^------------^----Do not want these spaces!
%
% a space would be appended to the last name and could cause every name on that
% line to be shifted left slightly. This is one of those "LaTeX things". For
% instance, "\textbf{A} \textbf{B}" will typeset as "A B" not "AB". To get
% "AB" then you have to do: "\textbf{A}\textbf{B}"
% \thanks is no different in this regard, so shield the last } of each \thanks
% that ends a line with a % and do not let a space in before the next \thanks.
% Spaces after \IEEEmembership other than the last one are OK (and needed) as
% you are supposed to have spaces between the names. For what it is worth,
% this is a minor point as most people would not even notice if the said evil
% space somehow managed to creep in.

% The paper headers
\markboth{Journal of \LaTeX\ Class Files,~Vol.~14, No.~8, August~2015}%
{Shell \MakeLowercase{\textit{et al.}}: Bare Demo of IEEEtran.cls for Computer Society Journals}
% The only time the second header will appear is for the odd numbered pages
% after the title page when using the twoside option.
% 
% *** Note that you probably will NOT want to include the author's ***
% *** name in the headers of peer review papers.                   ***
% You can use \ifCLASSOPTIONpeerreview for conditional compilation here if
% you desire.

% The publisher's ID mark at the bottom of the page is less important with
% Computer Society journal papers as those publications place the marks
% outside of the main text columns and, therefore, unlike regular IEEE
% journals, the available text space is not reduced by their presence.
% If you want to put a publisher's ID mark on the page you can do it like
% this:
%\IEEEpubid{0000--0000/00\$00.00~\copyright~2015 IEEE}
% or like this to get the Computer Society new two part style.
%\IEEEpubid{\makebox[\columnwidth]{\hfill 0000--0000/00/\$00.00~\copyright~2015 IEEE}%
%\hspace{\columnsep}\makebox[\columnwidth]{Published by the IEEE Computer Society\hfill}}
% Remember, if you use this you must call \IEEEpubidadjcol in the second
% column for its text to clear the IEEEpubid mark (Computer Society jorunal
% papers don't need this extra clearance.)

% use for special paper notices
%\IEEEspecialpapernotice{(Invited Paper)}

% for Computer Society papers, we must declare the abstract and index terms
% PRIOR to the title within the \IEEEtitleabstractindextext IEEEtran
% command as these need to go into the title area created by \maketitle.
% As a general rule, do not put math, special symbols or citations
% in the abstract or keywords.
\IEEEtitleabstractindextext{%
\begin{abstract}
As more and more Arabic texts emerged on the Internet, extracting important information from these Arabic texts is especially useful. 
As a fundamental technology, 
Named entity recognition (NER) serves as the core component in information extraction technology, while also playing a critical role in 
many other Natural Language Processing (NLP) systems, such as question answering and knowledge graph building. 
In this paper, we provide a comprehensive review of the development of Arabic NER, especially the recent advances in deep learning and pre-trained language model. Specifically, we first introduce the background of Arabic NER, including the characteristics of Arabic and existing resources for Arabic NER. Then, we systematically review the development of Arabic NER methods. Traditional Arabic NER systems focus on feature engineering and designing domain-specific rules. In recent years, deep learning methods achieve significant progress by representing texts via continuous vector representations. With the growth of pre-trained language model, Arabic NER yields better performance. Finally, we conclude the method gap between Arabic NER and NER methods from other languages, which helps outline future directions for Arabic NER.

\end{abstract}

% Note that keywords are not normally used for peerreview papers.
}

% make the title area
\maketitle

% To allow for easy dual compilation without having to reenter the
% abstract/keywords data, the \IEEEtitleabstractindextext text will
% not be used in maketitle, but will appear (i.e., to be "transported")
% here as \IEEEdisplaynontitleabstractindextext when the compsoc 
% or transmag modes are not selected <OR> if conference mode is selected 
% - because all conference papers position the abstract like regular
% papers do.
\IEEEdisplaynontitleabstractindextext
% \IEEEdisplaynontitleabstractindextext has no effect when using
% compsoc or transmag under a non-conference mode.

% For peer review papers, you can put extra information on the cover
% page as needed:
% \ifCLASSOPTIONpeerreview
% \begin{center} \bfseries EDICS Category: 3-BBND \end{center}
% \fi
%
% For peerreview papers, this IEEEtran command inserts a page break and
% creates the second title. It will be ignored for other modes.
\IEEEpeerreviewmaketitle

\IEEEraisesectionheading{\section{Introduction}\label{sec:introduction}}
% Computer Society journal (but not conference!) papers do something unusual
% with the very first section heading (almost always called "Introduction").
% They place it ABOVE the main text! IEEEtran.cls does not automatically do
% this for you, but you can achieve this effect with the provided
% \IEEEraisesectionheading{} command. Note the need to keep any \label that
% is to refer to the section immediately after \section in the above as
% \IEEEraisesectionheading puts \section within a raised box.

% The very first letter is a 2 line initial drop letter followed
% by the rest of the first word in caps (small caps for compsoc).
% 
% form to use if the first word consists of a single letter:
% \IEEEPARstart{A}{demo} file is ....
% 
% form to use if you need the single drop letter followed by
% normal text (unknown if ever used by the IEEE):
% \IEEEPARstart{A}{}demo file is ....
% 
% Some journals put the first two words in caps:
% \IEEEPARstart{T}{his demo} file is ....
% 
% Here we have the typical use of a "T" for an initial drop letter
% and "HIS" in caps to complete the first word.

\IEEEPARstart{A}{rabic} is the official language in the Arab world and one of the richest natural languages in the world in terms of morphological inflection and derivation. It is among top 10 most popular languages in the world with 420M native speakers, and more than 25 popular dialect \cite{guellil2021arabic}.
Arabic has many unique characteristics that make understanding of Arabic text both challenging and interesting. Thanks to the increasing availability of social media platforms, the amount of Arabic texts available on the Internet has seen a significant leap. 
 To effectively process these texts, Arabic Named Entity Recognition (ANER) attracts increasing attention. Named entity recognition (NER) is the task of identifying mentions that correspond to specific types, such as person name, location, and organization. Besides general domains, in several specific domains, such as the medical field, drugs, and clinical procedures can also be extracted by NER. With Arabic NER, it is convenient to extract important information in different domains.
NER can be used for different downstream tasks,
such as relation extraction\cite{cheng2021hacred}, entity linking \cite{gu2021read}, event extraction \cite{zhu2021efficient}, co-reference resolution \cite{clark2016improving}, and machine translation \cite{ugawa2018neural}.

% Thus, it is important to highlight recent advances in Arabic named entity recognition, especially recent deep learning-based NER architectures. 

As a basic component of natural language processing (NLP), NER is well-studied in English but still lacks comprehensive exploration in Arabic. What makes English NER easier than Arabic NER is that most of the mentions in English begin with capital letters which is not an option in Arabic. 
{In addition, Arabic is a morphologically complex language due to its extremely inflectional and the existence of many variations \cite{benajiba2008arabica}}. 
% There are Modern Standard Arabic (MSA) and Dialectal Arabic (DA), where each region even has its own dialect. 
There are three categories: Classical Arabic (CA), Modern Standard Arabic
(MSA) and Dialects Arabic (DA) \cite{elgibali2005investigating}. CA is the original Arabic language that has been used for over 1,500 years,
and it is usually used in most Arabic religious texts. MSA is one of the official languages of the United Nations and widely used in today’s Arabic newspapers, letters, and formal meetings, which is also focused by researchers. DA is spoken Arabic used in informal daily communication.

\begin{table}[t]
\centering
\begin{tabular}{l*{4}{c}}
\toprule

\textbf{Cite} & \textbf{Year} & \textbf{Type} & \textbf{Title} \\ \hline
\midrule
 \cite{shaalan2009nera} & 2009  &  Report &  \makecell[c]{NERA: Named entity recognition \\ for Arabic}        \\ \hline
\cite{shaalan2014survey} & 2014 & \textbf{Survey} & \makecell[c]{A survey of Arabic named entity \\ recognition and classification} \\ \hline

\cite{zirikly2015named} & 2015 & Report & \makecell[c]{Named entity recognition \\ for arabic social media} \\ \hline

\cite{dandashi2016arabic} & 2016 & \textbf{Survey} & \makecell[c]{Arabic named entity recognition \\ —a survey and analysis} \\ \hline

\cite{salah2017comparative} & 2017 & \textbf{Survey} & \makecell[c]{A comparative review of machine  \\ learning for  Arabic named \\ entity recognition} \\ \hline

\cite{el2019arabic} & 2019 & Report & \makecell[c]{Arabic named entity recognition \\ using deep learning approach} \\ \hline

\cite{liu2019arabic} & 2019 & Report & \makecell[c]{Arabic named entity recognition: What \\ works and what’s next} \\ \hline

\cite{ali2020recent} & 2020 & \textbf{Survey} & \makecell[c]{A recent survey of arabic named entity \\ recognition on social media.} \\ \hline

\bottomrule
\end{tabular}
\caption{Existing surveys and reports for Arabic NER. ``Report" denotes a competition report or describes a detailed system implementation. ``Survey" means a comprehensive investigation of different methods.}
\label{tab:ablation}
\end{table}

\begin{figure*}[tb]
    \centering
    \includegraphics{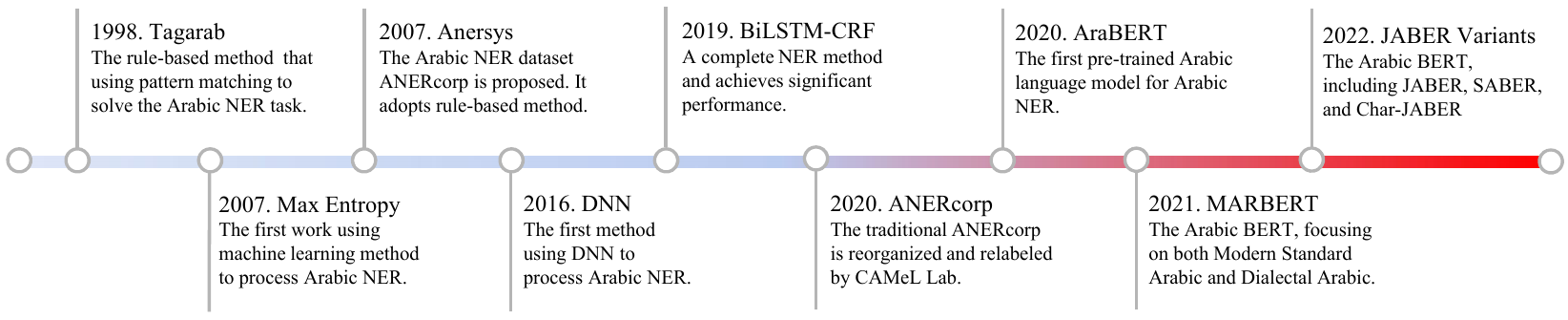}
    \caption{\text{Milestones in the development of Arabic NER.}}
    \label{fig:milesonte}
\end{figure*}

The development of solutions for Arabic NER has evolved according to advancements in algorithms and computational power. 
As to the techniques applied in Arabic NER, 
early NER systems depended on handcrafted rules. Afterward, machine learning approaches were trained on hand-crafted features to classify words into different named entity classes. Subsequently, deep learning techniques that rely on word embeddings and automatically discover representations for entity classification became dominant. More recently, the transformer-based pre-trained language model further promotes the development of this field with huge computational power. In this paper, we mainly focus on the development of named entities in Arabic. We do not claim this article to be exhaustive of all NER works on all languages.

\textbf{Motivation for conducting this survey.} Over the past few years, a large number of methods have explored deep learning methods in Arabic NER and achieved decent performance. We notice these methods adopt similar networks, mainly optional Bi-LSTM for encoding and CRF for decoding. However, there are much more abundant solutions in English and we hope these methods can also benefit the development of Arabic NER. Meanwhile, to the best of our knowledge, there are few systematic surveys in this field. As shown in Table 1, we list previous reports and surveys in Arabic NER. 
% Here we present reports 
Shaalan et al. \cite{shaalan2014survey} conducted a survey of Arabic NER in 2014 but focused on handcrafted features. 
Dandashi et al. \cite{dandashi2016arabic} and Salah et al. \cite{salah2017comparative} mainly explore the machine learning method for Arabic NER.  
More recently, Ali et al. \cite{ali2020recent} studied the Arabic NER on social media in 2020 but performed experiments on English datasets. 
In summary, existing surveys do not completely cover modern DL-based NER systems.
In Table 1, there are several reports introducing the competition method or system architecture, such as \cite{liu2019arabic} which is the champion of Arabic Named Entity Recognition challenge run by Topcoder.com. 
These reports describe an effective system but do not provide a comprehensive analysis of the development of Arabic NER.
There are also recent NER surveys focusing on deep learning technology \cite{li2020survey,yadav2018survey} in general domains in English. These surveys do not consider the language characteristic and specific challenges in Arabic.  
Thus, a comprehensive review of existing NER methods in Arabic is missing. We hope this review can help sort out the past Arabic NER methods systematically (as shown in Figure 1). It is worth noticing that discovering valuable information in a less rich language, such as Arabic is also a significant chance for the NLP community.

\textbf{Contribution for conducting this survey.} We summarize the contributions as below:
\begin{itemize}

\item To the best of our knowledge, this is the first comprehensive survey focusing on recent advances in Arabic NER, including the Arabic NER corpus, current resources, rule-based methods, machine learning methods, deep learning methods, and up-to-date pre-trained language model methods. In addition, we also survey the most representative methods for applied deep learning techniques in Arabic NER problems.

\item In this survey, due to the unfair comparison of previous methods from different data split, we carefully summarize the results of Arabic NER on the ANERCorp and AQMAR datasets. In this way, the researchers can easily compare different methods. We believe the fair comparison will contribute to more competitive Arabic NER methods.   

\item We outline the potential directions for Arabic NER,  including the alternative solutions for Arabic NER, fine-grained and nested Arabic NER, low-resource and distantly-supervised NER, and cross-linguistic knowledge transfer for Arabic NER. All these future trends will lead to more effective downstream applications. 

\end{itemize}

\begin{figure}[h!]
    \centering
    \includegraphics[scale=0.5]{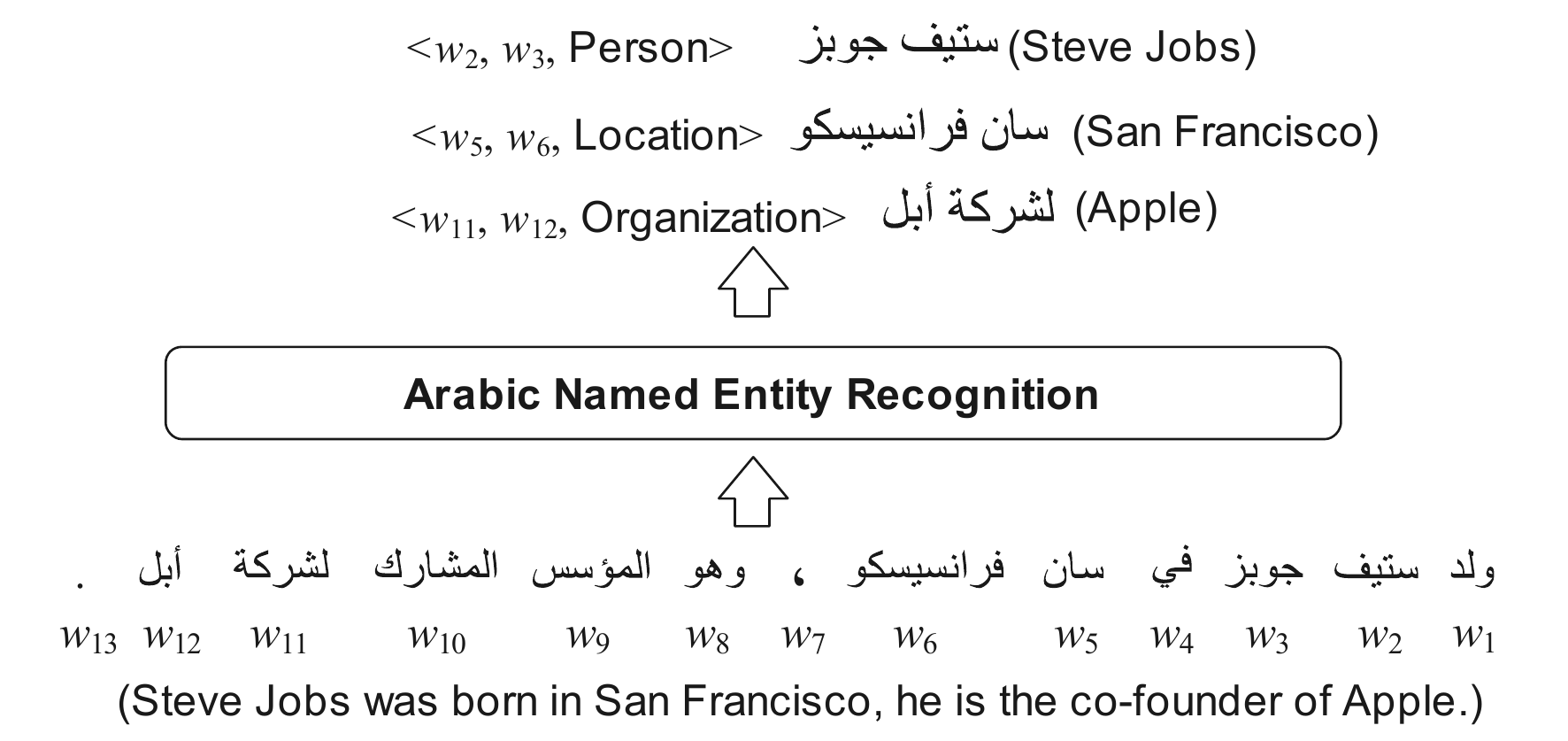}
    \caption{An illustration of the Arabic NER task.}
    \label{fig:what is arabic ner}
\end{figure}

% for IEEE Computer Society journal papers produced under \LaTeX\ using
% IEEEtran.cls version 1.8b and later.
% You must have at least 2 lines in the paragraph with the drop letter
% (should never be an issue)
% I wish you the best of success.

% \hfill mds
 
% \hfill August 26, 2015

% \subsection{Subsection Heading Here}
% Subsection text here.

% needed in second column of first page if using \IEEEpubid
%\IEEEpubidadjcol

\section{Background}
In this section, we first give a formal definition of Arabic NER. Subsequently, we introduce the challenges of Arabic NER.
Finally, we collect the current resources for Arabic NER and describe the evaluation metrics.

\subsection{What is Arabic NER}

\begin{figure}[t!]
\centering
\includegraphics[width=0.8\columnwidth]{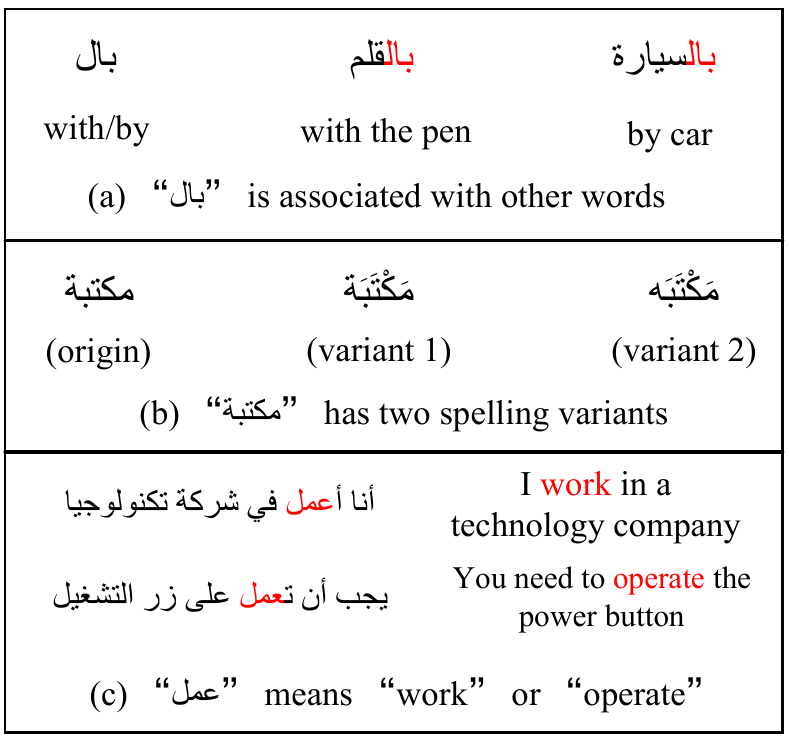}
\caption{Three examples of the challenges of Arabic NER.}
\label{fig:title_fig}
% \vspace{-4mm}
\end{figure}

Arabic NER aims to locate and classify named entities in Arabic text $s=<w_n, ..., w_2, w_1>$ (Different from the left-to-right style of writing in Chinese or English, Arabic is writing from right to left) into predefined entity categories. Same as other common NER tasks, the Arabic NER can also be divided into general domain including Person, Location and Organization types, and specific domain (e.g. Biomedical, News and Finance).

There are two typical outputs for NER, one is a list of tuples $<P_s, P_e, t>$, each of which is a mention of named entity. 
Here $P_s$ and $P_e$ are the start and end positions of a named entity mention, $t$ is the type of the mention from a predefined category set. Another one considers NER as a sequence labeling task, where each token is assigned to a special label (e.g. B-PER, I-PER, O). Figure \ref{fig:what is arabic ner} shows an example that an Arabic NER model recognizes three named entities from the given Arabic text. 
% \textbf{Different from English or Chinese NER, the complexities of Arabic often makes the Arabic NER task extremely difficult. The most intuitive concern is the various form including CA, MSA and DA. For a simple example, the word "telephone" borrowed from English in MSA is "هاتف" (hātif) and the word might be "تلفون" (telfoon) in DA, while in CA , it does not have a term for "telephone" because the concept did not exist during that time. Besides, other challenges of Arabic NER will be the Obstacles to the development of this task, they will be detailed described in the next subsection.}

\subsection{The Challenges of Arabic NER}
NER is especially challenging when it comes to Arabic because of its particularities and unique nature. The main characteristics of Arabic that pose non-trivial challenges for NER tasks are as follows: 
\begin{itemize}
\item \textbf{No Capitalization}: 
Capitalization is a very distinguishing orthographic feature, such as proper names and abbreviations, which helps NER systems accurately recognize specific types of entities. Unlike English, where a named entity usually begins with a capital letter, capitalization is not a feature of the Arabic words. Consequently, only relying on looking up the items in the lexicon of proper nouns is not an appropriate solution to this problem\cite{algahtani2012arabic}.

\item \textbf{The Agglutinative Nature}: 
    Arabic has a high agglutinative nature in which a word may consist of prefixes, lemma, and suffixes in different combinations, and that leads to a very complicated morphology\cite{abdelrahman2010integrated}. Meanwhile, named entities in Arabic are often associated with some other words, such as prepositions (e.g. for/to, as and by/with) or conjunctions (e.g. and)\cite{shaalan2014survey}. {As shown in Figure 3(a), the preposition is coupled with other words.}
% For example, the prepositions "\RL{بال}" (\textit{by/with}) in \RL{بالقلم } (\textit{with the pen}) and "\RL{بالسيارة }" (\textit{by car}).
% \textarabic{بال}
% {For example, the prepositions "بـ" (bi, \textit{by/with}) in “بالقلم ” (\textit{with the pen}) and "بالسيارة " (\textit{by car}) will make the NER system confused. Also, the analysis of the fragment "القاهرة والإسكندرية" (\textit{Cairo and Alexander}) that contains a conjunctions "و" yields "الإسكندرية" (\textit{Alexander}) as a location name.} 
A common treatment is to obtain the text segmentation separated by a space character\cite{benajiba2007anersys2}.
% such as the word xx consists of the named entity xx and a conjunction xx, we would separate it by a space character to xx.

\item \textbf{Spelling Variants}: In Arabic, the word may be spelled differently and still refers to the same word with the same meaning, creating a many-to-one ambiguity, which is also called transcriptional mbiguity\cite{shaalan2007person}, {as depicted in Figure 3(b)}. 
% \textbf{For instance, the word "مكتبة" (\textit{library}) can be spelled with diacritics as "مَكْتَبَة" or "مَكْتَبَه" and the word "مدرسة" (\textit{school}) can be written as "مَدْرَسَة" or "مَدْرَسَه" with the diacritics.}
% For example, there are some variants such as xx, xx, xx and xx when translating the named entity \textit{the city of Washington} from English to Arabic. 
To tackle this problem, researchers may normalize each variant to a unified form by a mechanism such as string distance calculation \cite{steinberger2012survey}.

\item \textbf{No Short Vowels}: Short vowels, or diacritics, are needed for pronunciation and disambiguation. However, most modern Arabic texts do not include diacritics. Therefore, one word form in Arabic may refer to two or more different words or meanings according to the context they appear, leading to a one-to-many ambiguity. {As illustrated in Figure 3(c), this word presents different meanings depending on the context in which it is used}. This phenomenon is easy to understand for a native Arabic speaker, but not for an Arabic NER system\cite{alkharashi2009person}. 
% \textbf{For example, the word "جمال" can mean either "camel" in "ركبتُ الجمل لأول مرة في صحراء مصر" (\textit{I rode a camel for the first time in the Egyptian desert}) or "beauty" in "أعجبتني جمال الطبيعة في هذه المنطقة" (\textit{I was impressed by the beauty of nature in this region}) depending on the context in which it is used.} 
Short vowels are very important because different diacritics represent different meanings, which could help to disambiguate the structure and lexical types of Arabic entities. 
For a long time, these ambiguities caused by short vowels can only be solved by using contextual information and some knowledge of the language\cite{benajiba2009arabic}. Recently, with the introduction of the Arabic pre-trained models (e.g. AraBERT \cite{antoun2020arabert} and AraELECTRA \cite{antoun2021araelectra191}), the challenge caused by diacritics is further alleviated.

% \item \textbf{Different usage of Arabic}: 
% Distinguished from languages such as English and Chinese, different usages of Arabic have significant impacts on named entity recognition. Arabic can be commonly classified into three categories: Classical Arabic (CA), Modern Standard Arabic (MSA) and Dialects Arabic (DA) \cite{elgibali2005investigating}. 
% CA is the original Arabic language that has been used for over 1,500 years, and it is usually used in most Arabic religious texts. MSA is one of the official languages of the United Nations and widely used in today’s Arabic newspapers, letters and formal meetings, which is also be focused by researchers. DA is the spoken Arabic used in informal daily communication.
% For this reason, it is significant to confirm the difference between various uses of the Arabic language.

\item \textbf{Lacking Linguistic Resources}: 
Adequate data resources, including annotated corpus and high-quality lexicons/gazetteers, are critical to building a strong NER system. However, unlike the languages such as English and Chinese that have large-scaled tagged corpus, {there is a limitation in the number of available Arabic linguistic resources for Arabic NER due to the absence of annotations \cite{oudah2012pipeline}}. Even so, a few Arabic corpora have been created and publicly available for research purposes, we will introduce more details of these Arabic resources in the section 2.3.

\end{itemize}

\begin{table*}[t]
	\centering
	\caption{List of annotated datasets for Arabic NER. ``\#Tags'' refers to the number of entity types.}	
	\label{tab:datasets}
	\begin{tabular}{ccccp{8.2cm}}
		\toprule
		Corpus & Year & Text Source & \#Tags & URL \\ \midrule
		
    ANERcorp & 2007     &  Website, news, magazines           & 4       & \url{https://camel.abudhabi.nyu.edu/anercorp/}    \\
    ACE 2003 &  2004    &  	Broadcast News, newswire genres     & 7       & \url{https://catalog.ldc.upenn.edu/LDC2004T09} \\

    ACE 2004 &  2004    &  Broadcast News, newswire genres             & 7       & \url{https://catalog.ldc.upenn.edu/LDC2005T09} \\

    ACE 2005 &  2005    &  Transcripts,  news            & 7       & \url{https://catalog.ldc.upenn.edu/LDC2006T06} \\

    ACE 2007 &  2007    &  Transcripts,  news            & 7       & \url{https://catalog.ldc.upenn.edu/LDC2014T18} \\

    REFLEX &  2009    &  Reuters  news           & 4       & \url{https://catalog.ldc.upenn.edu/LDC2009T11}    \\

    AQMAR &  2012    &  Arabic Wikipedia            & 7       &   \url{http://www.cs.cmu.edu/~ark/ArabicNER/}  \\

    OntoNotes 5.0 &  2013    &  News            & 18       & \url{https://catalog.ldc.upenn.edu/LDC2013T19}    \\

    WDC &  2014    &  Wikipedia          &  4       & \url{https://github.com/Maha-J-Althobaiti/Arabic_NER_Wiki-Corpus}    \\

    DAWT & 2017    &  Wikipedia           & 4    & \url{https://github.com/klout/opendata/tree/master/wiki_annotation}    \\

    % NE3L &   1997  &     New York Times news         &   7     &    \url{http://catalog.elra.info/en-us/repository/browse/ELRA-W0078/}  \\
    CANERCorpus & 2018  & Religion  & 20 & https://github.com/RamziSalah/Classical-Arabic-Named-Entity-Recognition-Corpus
    \\      

    Wojood &  2022    &  Wikipedia        & 21     & \url{https://ontology.birzeit.edu/Wojood/}    \\

			 \bottomrule
	\end{tabular}
\end{table*}

\begin{figure*}[tb]
\centering
\includegraphics[scale=0.7]{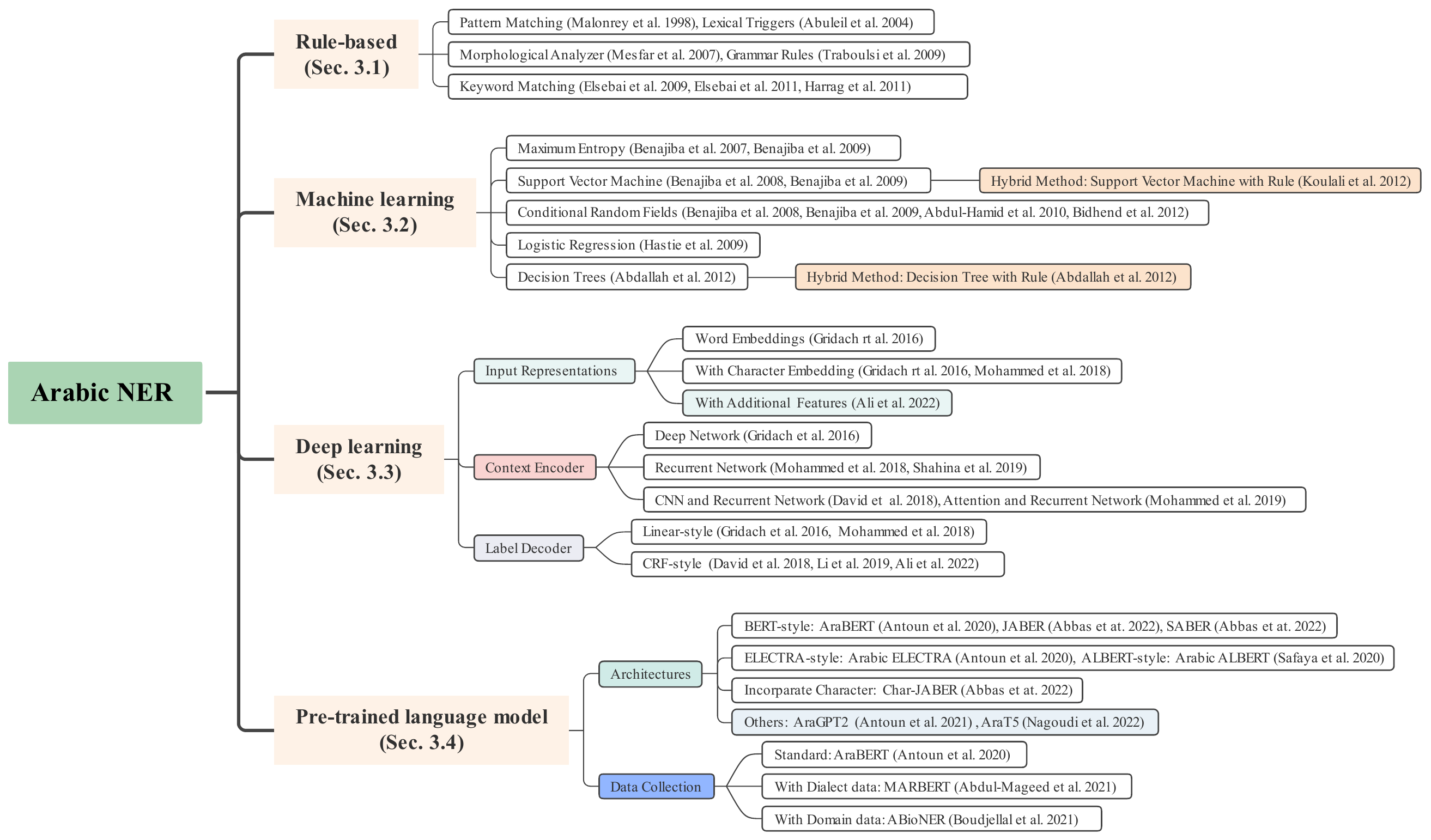}
\caption{The taxonomy of Arabic NER methods based on different methods. The rule-based method and machine learning method will be introduced in Sections 3.1 and 3.2. The methods of deep learning and pre-trained language model for Arabic NER will be described in Sections 3.3 and 3.4. In deep learning methods, we present the development of Arabic NER from three perspectives, including \text{input representations, context encoder, and label decoder}. In the pre-trained language model, we illustrate the Arabic NER methods according to their architectures and data collection scope.}
\label{fig:aner_deep_learning}
% \vspace{-4mm}
\end{figure*}

\subsection{Current Resources for Arabic NER}

In this section, we summarize some widely-used datasets and popular tools for Arabic NER. As shown in Table 2, most datasets are mainly developed by annotating news articles with a small number of entity types. {In consideration of the fact that certain Arabic NER corpora are exclusively accessible to individuals with paid license agreements, e.g., ACE 2003. It is difficult for researchers to access them as they are not free. Here we describe three representative datasets as below:} 
% As most works of Arabic NER report their results on ANER and AQMAR, here we also give a short description of these two datasets.
% As most works of Arabic NER report their results on ANER, here we also give a short description of this dataset.

ANERcorp\cite{benajiba2007anersys} consists of 316 articles from 11 diverse sources to acquire a generalized corpus and \text{is composed of a training set and a test set}. It contains 150,286 tokens. In addition to the traditional entities such as organization, location, and person, ANERcorp introduces the miscellaneous type to suggest a single class for entities that do not belong to any of the other classes. 
{It is worth noting that this dataset was split into many different configurations over the years which made it hard to compare fairly across papers and systems \cite{abdul2010simplified,zaghouani2012renar,morsi2013studying}}. In 2020, a group of researchers from CAMeL Lab met with the creator of the corpus to consult with him and collectively agree on an exact split, and accepted minor corrections from the original dataset. Thus, a new version is ANERcorp \footnote{https://camel.abudhabi.nyu.edu/anercorp/} is released and we suggest using this split as a standard dataset for evaluating NER performance in Arabic. 
% Different from the previous version, this new version does not have a development set. 
In this new version of the dataset, 
the sentences containing the first 5/6 of the words go to train and the rest go to test. The train split has 125,102 words and the test split has 25,008 words. 

{AQMAR\cite{mohit2012recall} is a small NER dataset based on Arabic Wikipedia which contains about 3,000 sentences. The creators picked 31 articles from history, technology, science, and sports areas that are suitable on the basis of length, cross-lingual linkages, and subjective judgments of quality. Compared with news text, the categories of Wikipedia articles are much more diverse. Accordingly, the traditional person, location and organization classes may not be suitable. And the appropriate entity classes can be quite different by domain. So AQMAR especially designs the custom entity categories for certain domains besides the classic ones.}

{CANERCorpus\cite{salah2018building} is proposed to support the studies of Classical Arabic  which is the official language of Quran and Hadith. Based on Islamic topics, the corpus contains more than 7000 Hadiths which covers all aspects of Prophet Muhammad's life and offers proper guidance to Islam. The named entity classes are especially designed for Islamic topics as well which covers Allah, Prophet, etc., which has not been handled before. Three experts with more than 5 years of experience in the field of computational linguistics completed the annotation. This work contributes a lot to the development of Classical Arabic.}

{Among these Arabic NER datasets, ANERcorp is the first to publish and followed by the most work, which represents the traditional Arabic NER. Besides the classic entity classes, AQMAR further incorporates domain entities in fields of history, technology, science and sports, which enables the studies of domain specific named entity recognition. However, compared with ANERcorp, it is much smaller. Before the CANERCorpus, there is no dataset for Classical Arabic. This dataset concentrates on the CA and annotates on large-scale standard Islamic corpus. Although the entity categories are also abundant, it is not so balanced. The propotion of some categories is even lower than 1\%.}

Besides the dataset, here we also introduce some commonly used Arabic NLP tools which contribute to Arabic NER processing. 
{Different from the end-to-end Arabic NER models that directly take the Arabic tokens as input and output the NER labels, in the traditional feature-based era, there are some important preprocess steps for Arabic NER, such as tokenization and part-of-speech tagging.}
MADAMIRA\cite{pasha2014madamira} \footnote{https://camel.abudhabi.nyu.edu/madamira/} provides {an online web demo for} morphological analysis and disambiguation of Arabic. It is designed to deal with basic tasks which are often the first or intermediate steps of a complex system, including diacritization, lemmatization, morphological disambiguation, part-of-speech tagging, stemming, glossing, and tokenization. The input text is converted into the Buckwalter representations and passed to the morphological analysis. Then the SVM extracts closed-class features while the language models acquire open-class features. The ranking component scores the analysis list of each word, which is subsequently passed to the tokenization component. Both the base phase chunking component and named entity recognizer take SVM to process the analyses and tokenizations.
{As shown in MADAMIRA\cite{pasha2014madamira}, the Arabic NER component directly benefits from the outputs of previous components.}

{Word segmentation is a key preprocess step in Arabic NER. 
However, most segmenters focus on Modern Standard Arabic, and perform poorly when dealing with Arabic text that contains dialectal vocabulary and grammar.
To process the dialectal Arabic text, Monroe et al. \cite{monroe-etal-2014-word} propose the Stanford Arabic segmenter that achieves 95.1\% F1 score on the Egyptian Arabic corpus.}
The Stanford Arabic segmenter \footnote{https://nlp.stanford.edu/projects/arabic.shtml} is an extension of the character-level conditional random field model. This segmenter expands the regular label space to introduce two Arabic-specific orthographic rules. 
Domain adaptation methods are applied to segment informal and dialectal Arabic text.

Farasa\cite{abdelali-etal-2016-farasa} \footnote{https://farasa.qcri.org/} is a fast and accurate Arabic segmenter based on SVM which ranks possible segmentations of a word by adopting a variety of features and lexicons. Farasa supports segmentation, stemming, named entity recognition, part of speech tagging and diacritization. The experiments demonstrate that this tool performs much better on the information retrieval tasks than Stanford Arabic segmenter and MADAMIRA.

As most Arabic NLP tools suffer from a lack of flexibility, CAMeL\cite{obeid2020camel} \footnote{https://github.com/CAMeL-Lab/camel\_tools} is presented to solve the problem. It is implemented in Python 3 and provides a series of pre-processing steps, such as transliteration and orthographic normalization. As for the named entity recognition, CAMeL fine-tunes the AraBert and experiments show that this tool outperforms the CRF-based system. 
{In more detail, CAMeL achieves 83\% overall F1 score on the ANERcorp dataset, surpassing the CRF-based system by a margin of 4\% F1 score.
Especially in the Person category, CAMeL achieves a 14\% improvement in F1 score compared to the CRF-based model.}

\subsection{Arabic NER metrics}

The metrics we take for Arabic NER are either the exact-match ones or the relaxed-match ones. For exact-match metrics, we evaluate the approaches with precision, recall, and F-score. Both the boundaries and entity type must be correct if the predicted entity is regarded as accurate. The precision suggests the ability to extract correct entities while the recall measures the ability to find as many entities as possible. Accordingly, F-score represents the balance of presion and recall. However, the exact-match metrics are somewhat too strict. Some approaches are also evaluated using relaxed-match metrics. For example, MUC-6\cite{grishman-sundheim-1996-message} propose to evaluate the system through the entity type and ignore the boundaries as long as there is an overlap with ground truth boundaries. Since the relaxed-match metric is complicated and inconvenient to analyze the bad cases, they are still not widely used.

\section{Method}
To help reveal the development of Arabic NER, we present several representative methods. As shown in Figure 4, the Arabic NER methods are summarized as four paradigms, including the rule-based methods, the machine learning methods, deep learning methods, and recent pre-trained language model methods. In the following sections, we will describe the important methods in each paradigm.

\subsection{Rule-based}

Before machine learning and deep learning methods are widely applied in NER tasks, the rule-based approach is most commonly used. Rule-based NER systems \cite{abuleil2004extracting,mesfar2007named,shaalan2008arabic,shaalan2009nera,elsebai2011extracting,abdallah2012integrating} depend on manually compiled local grammatical rules extracted and verified by experts in linguistics. 
Gazetteers and lexicons (or dictionaries) are utilized as grammar rules in the context where the named entities appear. Basically, most of the words in gazetteers are used as trigger words to help identify a named entity within the given text.

There are only a few works that leverage rule-based methods in named entity recognition tasks in Arabic because encoding human professional knowledge requires a lot of work from expert linguists, and usually just works for a single language.
At a very early stage of Arabic NER, Maloney and Niv \cite{maloney1998tagarab} proposed a rule-based pattern-matching method for Arabic NER tasks named TAGARAB. This system supports the recognition of Name, Date, Time and Numeric entities by combining high-precision morphological analysis and a pattern matching engine. 

However, the TAGARAB only considers the morphology of the Arabic, which is a low-level rule feature. Hence Traboulsi et al. \cite{traboulsi2009arabic} discussed the use of methods based on the so-called local grammar formalism in untagged Arabic corpora, in conjunction with a finite state automata to recognize patterns of person names in Arabic news texts. The results show that considering all the common grammars can save a lot of time and effort. Besides, Elsebai et al. \cite{elsebairule} adopted a rule-based method making use of the Buckwalter Arabic Morphological Analyser (BAMA) \cite{buckwalter2004issues} as well as using a set of keywords to analyze the texts that probably include named entities of person name, which achieved an F-measure of 89\%.

More recently, Harrag et al. \cite{harrag2011extracting} proposed a finite state transducer-based entity extractor to identify key entities from prophetic narrations texts. Concretely, they first adopted finite state transducer to assign a conceptual label among a finite set of labels to capture different types of named entities. Then for each detected concept, the goal of the system is to fill a pattern that contains the event description. The model achieved a 52.00\% F-measure on a set of prophetic narrations texts.

To sum up, rule-based NER methods are easy to understand based on solid linguistic knowledge. However, the shortcomings of rule-based systems are also noticeable. First of all, any maintenance or updates required for these rule-based systems is labor-intensive and time-consuming. Secondly, the problem is compounded if the linguists with the required knowledge and background are not available.

\subsection{Machine Learning}

Generally speaking, machine learning-based NER methods \cite{oudah2012pipeline} utilize ML algorithms to learn entity tagging decisions. They treat the NER problem as a classification task and require the availability of a large annotated dataset. 

Aiming at tackling Arabic NER tasks based on machine learning methods, Benajiba et al. (\cite{benajiba2007anersys,benajiba2007anersys2,benajiba2008arabicb,benajiba2008arabica}) did a lot of valuable works in the early stage. They first adopted Maximum Entropy (ME) \cite{benajiba2007anersys} and developed an Arabic NER system, ANERsys 1.0. The features used by the system are lexical, contextual and gazetteers features. However, this system is hard to recognize entities with more than one word. Subsequently, the developed ANERsys 2.0 \cite{benajiba2007anersys2} was proposed.

In order to further improve the performance of the Arabic NER system, Benajiba et al. \cite{benajiba2008arabicb} considers more features such as morphological, POS-tags and so on. They tested the new NER system on ACE Corpora and ANERcorp and obtained the state-of-the-art performance when all the features are used. 
Conditional Random Fields (CRF) based method is a strong baseline machine learning method \cite{bidhendi2012extracting}. Benajiba et al. \cite{benajiba2008arabica} further used CRF for improving ANERsys based on ME approach. The results showed that using the CRF method, the model can achieve nearly two percentage points better performance than ANERsys 2.0.
Other machine learning technologies, such as Logistic Regression\cite{hastie2009elements} and Hidden Markov Models (HMM) are also wildly used in the Arabic hybrid NER system and could further improve the performance compared with rule-based systems.

Besides solely relying on machine learning algorithms, a few methods \cite{abdallah2012integrating, oudah2012pipeline,koulali2012contribution} propose to integrate the rule-based approach and machine learning-based approach together to create a hybrid system with the aim of enhancing the overall performance of Arabic NER tasks. Specifically, the rule-based approach relies on handcrafted grammatical rules, while ML-based approach takes advantage of the ML algorithms that utilize sets of features extracted from datasets for building NER systems. Then, the hybrid approach combines rule-based approach with ML-based approach together in a pipeline process to improve the overall performance of the system. For instance, 
Abdallah et al. \cite{abdallah2012integrating} combines the Decision Tree with rule to recognize three types of named entities including person, location, and organization, and proved its comparatively higher efficiency as a classifier in the hybrid NER system. 
It is worth noting that the hybrid approach suffers from an increase in complexity because it involves the use of grammatical rules and handcrafted features.
% Support Vector Machine (SVM) is an effective machine learning method for Arabic NER. 

Although the machine-learning methods achieve decent performance, they are not able to learn complex and high-level features from data through linear models such as log-linear HMM or linear chain CRF. To make up the deficiency of machine learning, deep learning methods are presented to discover hidden features automatically. We will introduce deep learning methods in section 3.3.

\subsection{Deep Learning}

Recently, deep learning methods have been widely proposed for Arabic NER. Compared to traditional handcraft feature based approaches, deep learning methods are able to learn more robust features and reduce the effort of designing handcraft features. Moreover, the deep learning methods can be trained in an end-to-end manner by gradient descent. Benefiting from this, we can 
design possibly more complex Arabic NER systems. In this section, as shown in Figure 4, we will introduce the development of deep learning methods in Arabic NER from three perspectives, namely the input representations, the context encoder, and the label decoder. The \textit{input representations} consider word- and character-level embeddings as well as additional features which are effective for Arabic NER. \textit{Context encoder} captures the context dependencies from the text using CNN, RNN, or other networks. \textit{Label decoder} aims to predict corresponding labels for tokens in the input sequence.

\begin{figure}[t!]
\centering
\includegraphics[width=0.8\columnwidth]{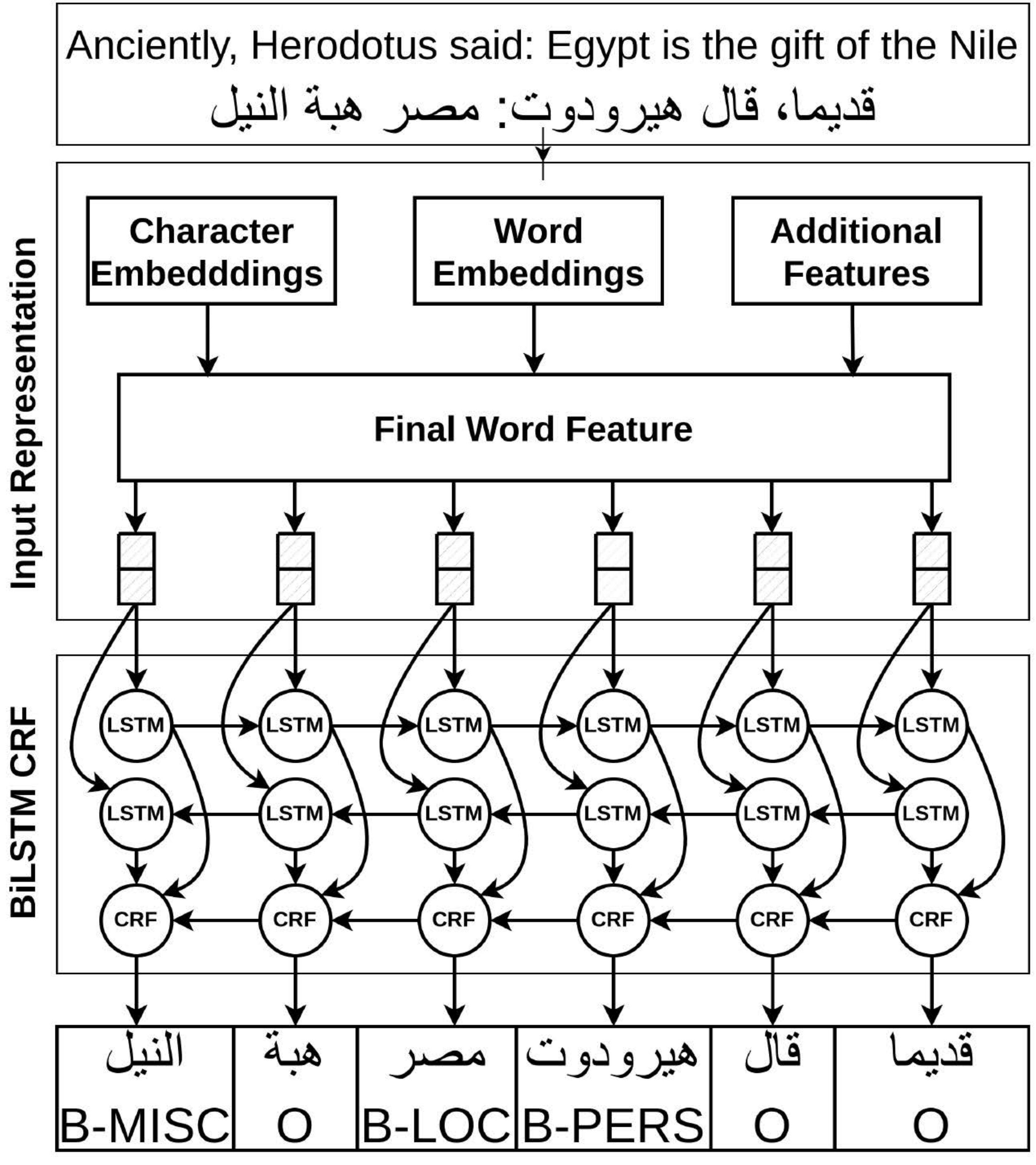}
\caption{Architecture of a representative deep learning method \cite{lotfyenhancing} for Arabic NER. This model is a typical BiLSTM-CRF architecture and adopts multiple input features.}
\label{fig:title_fig}
% \vspace{-4mm}
\end{figure}

To apply deep learning in Arabic NER, Gridach et al. \cite{gridach2016deep} first introduce a simple deep neural network (DNN) for Arabic NER in 2016. Based on the word2vec \cite{mikolov2013distributed} which represents words in low dimensional real-valued dense vectors, they further construct three neural network layers, including an input layer, a hidden layer, and an output layer. In the input layer, they explicitly represent context as a window consisting of a center word concatenated with its immediate surrounding neighbors. The hidden layer learns feature combinations from the word representations of the window. Finally, the output layer transforms the hidden representations and classifies them into different entity types. 
% This work explores word embedding as the input representation for neural networks. 
This work adopts a simple linear layer to predict the label sequence, while the following works mostly utilize conditional random fields to determine the label sequence. 

Subsequently, more complex context encoders are applied for Arabic NER, such as the Recurrent neural network (RNN). 
 RNN effectively captures the contextual information for each token and has achieved remarkable success in many different NLP tasks \cite{zhao2019recurrent,lin2021asrnn}.
In Arabic NER, Mohammed et al. \cite{fi10120123} adopted a bidirectional long short-term memory (BiLSTM) layer for entity recognition. The BiLSTM layer can efficiently use both
past and future input features \cite{huang2015bidirectional}, which makes it effective for the NER task. Specifically, they first apply a character-level embedding layer and a word embedding layer as model inputs. 
% However, word embedding treats words as the smallest unit and disregards any morphological resemblances between various words, thereby leading to the OOV problem. Meanwhile, character embedding can operate over individual characters in each word. 
Subsequently, they further propose an embedding attention layer to integrate these two kinds of embeddings. The integrated embeddings are then fed into the BiLSTM layers and the final classification layer. To evaluate the effectiveness of different recurrent networks, Shahina et al. \cite{shahina2019sequential} explore RNN, LSTM, and GRU to model the Arabic NER data where context of each word in the sequence is relevant. 

As to input representations for Arabic NER, 
% Gridach et al. \cite{gridach-2016-character} propose a neural-network-based model for recognizing named entities in Twitter.
Gridach et al. \cite{gridach-2016-character} investigate the effectiveness of injecting character-level features based on the commonly used BiLSTM-CRF architecture. Specifically, they employ a BiLSTM encoder to encode the characters and concatenate the first and last character representations as the final character-level representation for the corresponding word. This character representation is then concatenated with the word embedding as the input to the NER model encoder.
Besides, this paper shows that using pre-trained word embeddings outperforms randomly initialized word embeddings by a large margin. Finally, 
Gridach et al. \cite{gridach-2016-character} evaluate the proposed model on the Twitter NER dataset \cite{darwish2013named} and the results show the effectiveness of introducing character-level information and pre-trained word embeddings.

To encode both word-level and character-level input representations, David et al. \cite{awad2018arabic} also adopt the BiLSTM to encode the combination of character- and word-level representations. The word embedding is passed into a CNN layer before being fed into the BiLSTM. 
% For tagging decoding, they use Conditional Random Field (CRF) instead of normal classification. 
Mohammed et al. \cite{ali2019boosting} propose an efficient multi-attention layer system for the Arabic NER task. Similar to the previous work, they adopt  character-level embeddings and word embeddings. Differently, they further introduce a self-attention layer after encoding the embeddings with the BiLSTM, which can boost the performance of the system. 

Besides traditional character embeddings and word embeddings, as shown in Figure 5, Ali et al. \cite{lotfyenhancing} explore additional word features from  an external knowledge source. The basic architecture is also BiLSTM-CRF. They add morphological and syntactic features to different word embeddings to train the model. Specifically, they use the 2019 release of the Madamira tool \cite{pasha2014madamira} to generate the Part-of-Speech (POS), capitalization, and word analysis features. 
Furthermore, they design a fourth feature which is the quote feature. All features are represented as one-hot vectors and concatenated to the word and character embeddings.

Besides supervised training, Helwel et al. \cite{helwe2019arabic} propose a semi-supervised co-training approach to the realm of deep
learning, named deep co-learning. Specifically, they use a small amount of labeled data with a large amount of partially labeled data from Wikipedia. They first train two deep learning classifiers based on the fully-labeled dataset and different sets of word embeddings. These two models serve as different views for partially labeled data. Subsequently, they iteratively retrain classifiers by applying each classifier on the partially labeled corpus and using the output of each classifier to re-train the other classifier.

\begin{figure*}[tb]
    \centering
    \includegraphics{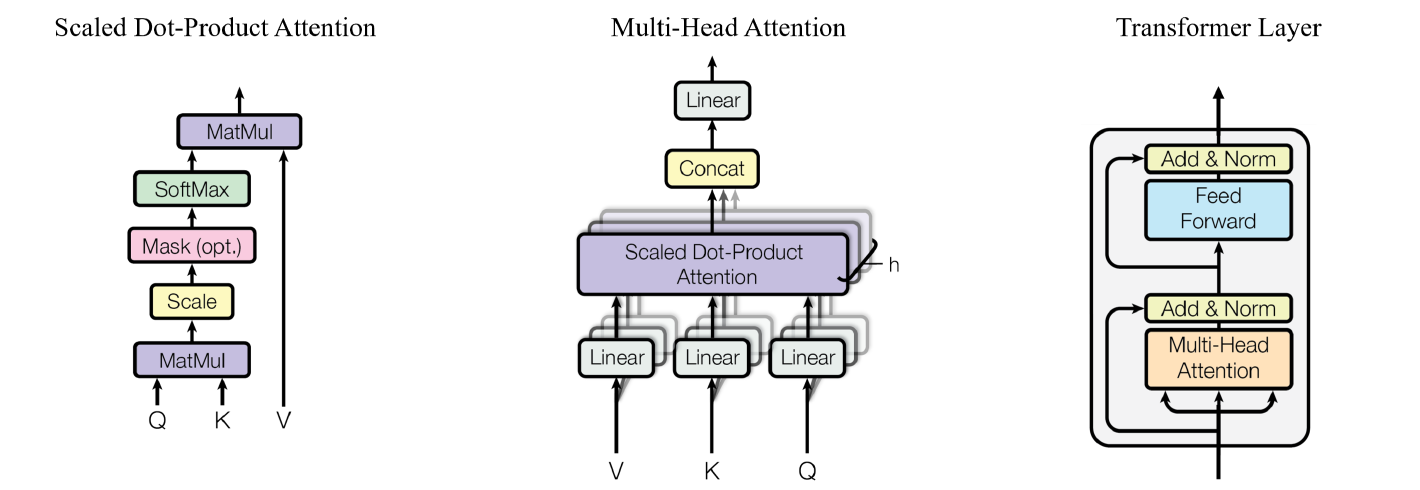}
    \caption{Overview of scaled dot-product attention (left), multi-head attention (middle), and Transformer layer (right). The figure is from \cite{vaswani2017attention}.}
    \label{fig:milesonte}
\end{figure*}

Liu et al. \cite{liu-etal-2019-arabic} propose an integrated Arabic NER system that combines deep representation learning, feature engineering, sequence labeling, and model ensemble.
Following the common practice, Liu et al. \cite{liu-etal-2019-arabic} treat Arabic NER as a sequence labeling problem and employ the BiLSTM-CRF architecture as one baseline.
To boost the model performance, Liu et al. \cite{liu-etal-2019-arabic} introduce a dictionary-based method to enhance the model outputs.
Specifically, they first build a named entity map from the surface words to their entity types.
The map will be used in the prediction layer that adds extra entities if the surface words match some entities in the map and there is no conflict between the surface words and the model predictions.
Experiments show that the dictionary-based method can improve the recall value at a relatively small cost of the precision value, therefore improving the F1 score.
Besides, handcrafted features are also employed in this work that categorize the input tokens into three types: numbers, English words, and Arabic words and are mapped into vectors.
Experimental results show a significant improvement of 4 F1 score on the AQMAR dataset.

Compared with rule-based and machine learning based NER methods, deep-learning-based Arabic NER shows its superior performance over the past decade.
In addition, deep-learning-based Arabic NER does not require hand-crafted features, which simplifies the usage of Arabic NER.

\subsection{Pre-trained Language Model}

% This file describes the recent language models and transformer architectures methods.
% \begin{figure}[!tb]
%     \centering
%     \includegraphics[scale=0.5]{Figures/BERT.pdf}
%     \caption{Model architecture of BERT.}
%     \label{fig:BERT}
% \end{figure}

% \begin{figure}
%     \centering
%     \includegraphics[scale=0.4]{Figures/transformer.pdf}
%     \caption{Architecture of the transformer block.}
%     \label{fig:transformer}
% \end{figure}

In recent years, pre-trained language models (PLMs) have shown superior representative abilities to capture contextual information in sentences, thus achieving significant improvements in a variety of NLP tasks, including NER.
Generally, most PLMs employ the Transformer architecture as the encoder, which consists of two layers, i.e., a multi-head attention layer and a 2-layer feed-forward neural network.
Figure 6 (right) shows the architecture of one transformer block, where Figure 6 (middle) shows the details of the multi-head attention \cite{vaswani2017attention} and Figure 6 (left) depicts the inner scaled dot-product attention. The most important key elements to achieve a strong PLM depend on efficient model architectures and high-quality data collection. Thus, we will describe the Arabic PLM and its applications to Arabic NER from these two perspectives.

BERT \cite{kenton2019bert} is one of the most famous pre-trained language models that uses masked language model (MLM) and next sentence prediction (NSP) as the learning objective. 
For MLM, a subset of the input tokens are masked and the objective is to recover the original token at the masked position.
The NSP task allows BERT to learn the relationship between two sentences by predicting whether the second sentence is the true next or not.
% \textcolor{red}{MLM NSP}
% Figure \ref{fig:BERT} shows the overall model architecture.
By pre-training on large-scale unlabeled datasets, BERT learns rich text information and can provide powerful contextual information that facilitates downstream tasks. 
% There are two commonly used strategies to use BERT for downstream tasks, i.e., fixed representation and fine-tuning.
% The fixed representation strategy usually extracts representations from several BERT transformer layer outputs (such as the last four layers) and treats them as external input representations for downstream tasks.
% Differently, the fine-tuning method usually treats the whole BERT model as the downstream NLP task encoder and additionally adds a task-specific decoder (such as a MLP classifier) on top of BERT for task prediction.
However, most of the previous PLM works focus on English, with little attention to the Arabic language.
In this section, we mainly review PLM works for Arabic, demonstrating the effectiveness of Arabic PLM for Arabic NER.

Antoun et al. \cite{antoun2020arabert} pre-trained an Arabic BERT named AraBERT for the purpose of achieving the same success as BERT for English.
Following BERT, Antoun et al. \cite{antoun2020arabert} also employ MLM and NSP as the training objective. 
% The used Arabic pre-trained data contains about 24GB text of news domain from different media in different Arab regions.
To accommodate the characteristic of Arabic language, the authors propose to segment words into stems, prefixes, and suffixes first with an offline segmenter.
The pre-trained model is evaluated on three Arabic language understanding tasks, including sentiment analysis, named entity recognition, and question answering.
% Specifically, AraBERT achieves new state-of-the-art of 84.2 F1 score on the ANERcorp dataset.

Antoun et al. \cite{antoun2021araelectra191} build AraELECTRA, whose model architecture is the same as ELECTRA \cite{Clark2020ELECTRAPT}.
The entire model contains two neural network modules, i.e., 1) a generator that takes the corrupted sentences as input and learns to predict the original tokens that have been masked and 2) a discriminator that takes as input the predicted sentence and try to tell which tokens are masked and which tokens are original. 
% The pre-trained model is trained on the same dataset as ARABERTV0.2 \cite{antoun2020arabert}, which contains about 77GB texts mainly from news domain.
In the fine-tuning stage, the input sentence is fed into the discriminator with an additional linear layer for downstream tasks.
They evaluated the Arabic ELECTRA  model on the ANERCorp dataset and the results reveal its superiority.

Besides training objectives, the lightweight pre-trained model also attracts increasing attention for a faster inference speed.  
Following English ALBERT \cite{lan2019albert}, Safaya et al. \cite{ali_safaya_2020_4718724} pre-trained an Arabic ALBERT with Arabic texts.
The main difference between BERT and ALBERT is that ALBERT has the same parameter across different layers, thus largely reducing the model parameter size. Considering the language characteristic of Arabic, Ghaddar et al. \cite{ghaddar2022revisiting} propose a Char-JABER which enhances the BERT-style Arabic PLM with character-level representations at the input layer. 

There are also some other style Arabic PLMs except the encoder-style AraBERT, such as decoder-style AraGPT2 \cite{Antoun2021AraGPT2PT} and encoder-decoder-style AraT5 \cite{nagoudi-etal-2022-arat5}.
These works are promising for the development of Arabic NLP.
However, since these models are not evaluated on Arabic NER tasks, we will not introduce them in detail.
% Please refer to their original paper if the readers are interested.
% According to some research in English, the generation-style PLMs may not be suitable for complicated structure prediction problems, like named entity recognition and semantic parsing.

% \begin{table}[!tb]
%     \centering
%     \begin{tabular}{c|c|c|c }
%     \hline
%         &Model archi. & Data & Results on ANER\\
%         \hline
%         AraBERT & BERT &24GB & 84.2 \\
%          AraELECTRA & ELECTRA&77GB &83.95 \\
%          Arabic-ALBERT & ALBERT& 4.4Billion &- \\
%          ARBERT &BERT &61GB &84.38 \\
%          MARBERT &BERT &128GB &80.64 \\
%     \hline
%     \end{tabular}
%     \caption{Summary of recent PLM works. Note that we report the used data size with different measuring unit, i.e., GB of storage size and number of tokens, according to these original works.}
%     \label{tab:plm}
% \end{table}

Based on the above proposed Arabic PLMs, Norah et al. \cite{Alsaaran2021ClassicalAN} propose the first work that tackles the classical Arabic NER. They propose to fine-tune a BiLSTM/BiGRU-CRF model on Arabic BERT and compare with two simple baselines, i.e., 1) a linear layer with softmax function on top of BERT (\texttt{BERT}) and 2) a CRF layer on top of BERT (\texttt{BERT-CRF}). They conduct experiments on the classic Arabic NER dataset CANERCorpus. The experimental results show that the two simple baselines (\texttt{BERT}, \texttt{BERT-CRF}) have already achieved promising results of 94.45 and 94.68 F1 score, respectively.
The BERT-BiGRU-CRF model achieves the best result of 94.76 F1 score.

To further explore the effectiveness of different PLMs architectures, Al-Qurishi et al. \cite{al-qurishi-souissi-2021-arabic} propose a Transformer-based CRF model for Arabic NER that fine-tuned on three BERT variants, i.e., AraBERT, AraELECTRA, and XLM-Roberta.
The experiments are conducted on the ANERCorp and AQMAR datasets and the results show that fine-tuning on AraBERT achieves the best performance. % of 91.35 F1 score.
% The proposed model used the IOB tagging schema

Apart from the fully supervised methods for Arabic NER, Helwe et al. \cite{helwe-etal-2020-semi} propose a semi-supervised framework based on a teacher-student learning mechanism, called deep co-learning.
% This work uses two datasets, i.e., one labelled dataset and one partially-labelled dataset.
% The labelled dataset contains samples that are annotated with NER labels on every token, while the sentences in the partially-labelled dataset are only annotated part of the tokens.
The basic Arabic NER backbone is based on the AraBERT.
In the training process, they first train a teacher model with the labelled dataset.
Then, the trained teacher model is used to predict the pseudo labels on the partially labelled dataset.
Finally, the result pseudo dataset is used to train a student model.
% Finally, the student model is fine-tuned on the labelled dataset and evaluated on the test data.
% The trained model achieves better results on AQMAR and NEWS datasets than the fully supervised model with a large margin of 4 F1 score.
However, the trained model has lower performance on the TWEETS dataset \cite{el2013kalimat} compared with the other two datasets, i.e., AQMAR and NEWs dataset \cite{el2013kalimat}.
The reason may come from that the AraBERT is trained mainly on the MSA corpus while the TWEETS dataset mostly belongs to the Egyptian dialect. 
This phenomenon highlights another important component  for Arabic NER, i.e., high-quality data collection.

% To handle this problem, Abdul-Mageed et al. \cite{abdul2021arbert} trained an Arabic PLM MARBERT with the collected tweets dataset and we will discuss it later. 

% In addition, they also compare the fine-tuning strategy with the feature-extraction strategy and the results prove that the fine-tuning method has a better performance. 

To extend the Arabic PLMs to diverse data domains and different Arabic dialects, Abdul-Mageed et al. \cite{abdul2021arbert} collect the training data from several sources, i.e., books, new articles, common crawl data, and Wikipedia and train an Arabic PLM MARBERT. 
% The total size of the collection is 61 GB and the authors perform light pre-
% processing to the data to ensure the faithfulness.
% Different from mBERT and XLM-R, this work uses a larger vocabulary of 100K WordPieces.
% Most of the training settings are the same as BERT, such as the model configuration, the whole word masking, and so on.
% There is an interesting setting that this work uses a fixed learning rate of 1e-4 for all of their models.
To promote the understanding of the Arabic dialect, they also train the model with a larger Twitter dataset.
The Twitter dataset is collected from a large in-house dataset of 6 billion tweets, resulting in a collection of 128GB texts.
In the training process, this work removes the next sentence prediction task since the tweets are usually short.
The extensive experiments perform on five NER datasets, i.e., ACE03NW, ACE03BN, ACE04NW, ANERcorp, and TW-NER.
We notice that MARBERT achieves a new SOTA on TW-NER, verifying the effectiveness of pre-training with tweets.
% In order to evaluate the trained models, this work also propose a benchmark that concatenating 42 tasks in the same category into 6 clusters, namely ARLUE.
% The MARBERT-v2 achieves the best overall score on ARLUE compared with several competitive PLMs, including AraBERT.
% To promote the development of Arabic PLM research, the authors also make their work open-sourced.

In addition to the commonly used news-domain NER dataset, several works also try to promote the progress of the Arabic NER in the biomedical domain.
% Biomedical Arabic NER is more challenging because of the lack of annotated resources and the complex characteristic of Arabic language.
Biomedical Arabic NER is more challenging due to its significant domain divergence. For this reason, the typical Arabic PLMs, such as AraBERT, usually perform not satisfactory. To solve this problem,    
Boudjellal et al. \cite{Boudjellal2021ABioNERAB} propose an Arabic biomedical pre-trained model by injecting biomedical knowledge into the AraBERT pre-trained model. They manually collect a set of literature corpora from medical journals, such as 
PubMed, MedlinePlus Health Information.
The collected corpus contains about 500k words, which is much smaller compared with the data used to train AraBERT. 
Even though, the experimental results show that the trained biomedical model achieves better performance than AraBERT on the bioNER data.
% Moreover, the trained ABioNER model shows its superiority with only few training steps,
This result also demonstrates the value of training a domain-specific pre-trained model for the corresponding domain-specific task.
% This work motivates us that if we need to focus on some specific domain, such as biomedical, financial, and so on, we only need to collect a few domain data and continue training with the base Arabic pre-trained model.
% The resulting model can learn the domain-specific knowledge and can thus have a strong ability to do related domain tasks.

\begin{table*}[t]
	\centering
	\caption{Comparison of Arabic NER systems on ANERCorp. The results are reported on four kinds of entities, including person,
location, organization and miscellaneous. 
% BiGRU means Bidirectional Gated Recurrent Unit. CRF indicates a conditional random field. 
``Standard" in the Dataset Split means the methods use the official dataset split instead of splitting the dataset by themselves. As traditional rule-based and machine learning methods do not have a specific context encoder, we merge columns ``Input representations" and ``Context encoder". ``-" symbols in the table indicate the corresponding item is not described in the original paper.}	
	\label{tab:tabsurvey}

 \scalebox{0.98}{

	\begin{tabular}{c|c|c|c|c|c}
		\hline \hline

{Work} & Year & {Input Representations} and {Context Encoder} & Label Decoder & Dataset Split & {Performance (F1 score)}                        \\  \hline  \hline

\cite{benajiba2007anersys} & 2007 & N-Gram & Maximum Entropy & Standard & 54.11 \\ \hline

\cite{benajiba2008arabica} & 2008 & - & CRF & Standard & 65.91 \\ \hline

\cite{abdul2010simplified} & 2010 & N-Gram & CRF & 4:0:1 & PER: 81, LOC: 88, ORG: 73 \\ \hline 

\cite{koulali2012contribution} & 2012 & - & SVM-HMMs & Standard & \makecell[c]{PER: 94.4, LOC: 96.9, \\ ORG: 80.66, MISC: 60.82} \\ \hline

\cite{zaghouani2012renar} & 2012 & - & Rule & Standard & 67.13 \\ \hline 

\cite{morsi2013studying} & 2013 & - & CRF & 3:0:1 & 68.05 \\ \hline

\cite{shaalan2014hybrid} & 2014 & - & Rule, Mechine Learning & Standard &  PER: 94, LOC: 90, ORG: 88 \\ \hline

\cite{dahan2015first} & 2015 & N-Gram & HMM & - & PER: 79, LOC: 78, ORG:67 \\ \hline

\cite{gridach2016deep} & 2016 & Word2vec, DNN & Linear & Standard & 88.64 \\ \hline 

\cite{mesmia2018casaner} & 2018 & - & Rule & Standard & PER: 78, LOC: 67, ORG: 63 \\ \hline

\cite{fi10120123} & 2018 & Word2vec, CNN-based Character & - & 8:1:1 & 87.12  \\ \hline

\cite{khalifa2019character} & 2019 & Word2vec, CNN-based Character, CNN, LSTM & CRF & 8:1:1 & 88.77   \\ \hline

\cite{elsherif2019arabic} & 2019 &  - & Rule & Standard & PER: 83, LOC: 92, ORG: 89 \\ \hline

\cite{el2019arabic} & 2019 & Bi-LSTM & CRF & - & 90.6 \\ \hline 

\cite{antoun2020arabert} & 2020 & AraBERT v1 & Linear & - & 81.9 \\ \hline

\cite{antoun2020arabert} & 2020 & AraBERT v0.1 & Linear & Standard & 84.2 \\ \hline

\cite{alsaaran2021arabic} & 2020  &  AraBERT v0.1, BiGRU &  Linear & 8:1:1 & 90.51 \\ \hline

\cite{obeid2020camel} & 2020 &  -  & - & 5:0:1  &  83 \\ \hline

\cite{rom2021supporting} & 2021 & - & - & 5:0:1 & 81.39 \\ \hline

\cite{antoun2021araelectra191} & 2021 & AraELECTRA & Linear & 5:0:1  &  83.95 \\

\hline

\cite{al-qurishi-souissi-2021-arabic} & 2021 & AraBertv2 & CRF & 5:0:1 & 89.56  

\\ \hline 

\cite{al-qurishi-souissi-2021-arabic} & 2021 & AraELECTRA & CRF & 5:0:1 & 87.21  

\\ \hline

\cite{ghaddar2022revisiting} & 2022 & MARBERT  & Linear & 5:0:1 & 80.5 \\ \hline

\cite{ghaddar2022revisiting} & 2022 & Arabic-BERT & Linear & 5:0:1 & 82.05 \\ \hline

\cite{ghaddar2022revisiting} & 2022 & ARBERT & Linear & 5:0:1 & 84.03 \\ \hline

\cite{ghaddar2022revisiting} & 2022 & JABER & Linear & 5:0:1 & 84.20 \\ \hline

\hline

	\end{tabular}}
\end{table*}

Up to now, Arabic NER has achieved significant progress with the surge of pre-trained language models.
On the one hand, existing Arabic PLMs provide powerful representations for Arabic texts that effectively boost the Arabic NER  performance.
On the other hand, more Arabic PLMs for specific domains are still urgently needed for Arabic NER from corresponding domains.
% different style PLMs offer different methods to handle the NER task, such as sequence labeling and generation.
% Besides, we hope Arabic named entities could also be used in PLMs to inject entity information for improving downstream Arabic tasks.

% \renewcommand\arraystretch{2}

% \begin{table*}[t]
% 	\centering
% 	\caption{Comparison of Arabic NER systems on AQMAR. The results are reported on three kinds of entities, including person, location, organization. BiGRU means Bidirectional Gated Recurrent Unit. CRF indicates a conditional random field.}	
% 	\label{tab:tabsurvey}

%  \scalebox{0.98}{

% % \setlength{\tabcolsep}{1 mm}{
% 	\begin{tabular}{c|c|c|c|c|c|c}
% 		\hline \hline

% {Work} & Year & {Input representations} & {Context encoder} & Tag decoder & Dataset Split & {Performance (F1 score)}                        \\  \hline  \hline

% \cite{pasha2014madamira} & 2014 & - & - & - & Standard & 29.2 \\ \hline

% \cite{abdelali-etal-2016-farasa} & 2016 & - & - & - & Standard & 52.9 \\ \hline

% \cite{helwe2020semi} & 2020 & AraBERT & - & Linear & Standard  & 61.5 \\ \hline 

% \hline

% 	\end{tabular}}
% \end{table*}

\subsection{Discussion}
After reviewing the four paradigms for Arabic NER, here we organize the experiment results on Arabic NER datasets in the order of publish date for further comparison. 
As shown in Table 3, we conclude the experimental results on ANERCorp. Due to the difference in data split strategies between methods, it is not easy to fairly compare them. In other words, the Arabic NER lacks a standard leaderboard, while the ANERCorp already has the training, development, and test set.

From this table, we can observe that most traditional methods, including rule-based methods and machine learning methods, perform significantly worse than deep learning methods. {These methods usually rely on specifically designed rules or N-grams, leading to weak generalization ability.
For deep learning methods, they mainly utilize LSTM for representation encoding and adopt CRF for tag decoding. LSTM effectively encodes the sentence information compared to traditional RNN. Meanwhile, CRF can achieve a more robust performance than a simple linear layer.}
{For input representations,} it is a common choice to encode the character representation by a CNN network and then integrate the character-level feature with the word-level feature.
{This strategy is important for encoding Arabic texts as character representation provides morphological and syntactic information. }  
When it comes to pre-trained language model, the model performance achieves a significant leap. As shown in Table 3, comparing different pretrained model AraELECTRA, MARBERT, Arabic- BERT, ARBERT, and JABER, we can observe that JABER obtains the best result 84.20 when using linear decoder. We believe this result can further improve as JABER uses a simple linear layer for label decoding. {However, it is worth noting that these pre-trained models are trained with different training corpus. In other words, the practical effectiveness depends on the downstream scenarios when applying them, especially when facing Arabic dialects.}

\subsection{Applied Techniques For Arabic NER}

% This file describes applications, such as multi-task learning, transfer learning for Arabic NER.

Previous sections outline typical network architectures for Arabic NER. In this section, we survey recently applied
techniques that are being explored for Arabic NER.

\subsubsection{Deep Multi-Task Learning for Arabic NER}

Multi-task learning\cite{ruder2017overview,zhang2018overview} has been wildly applied in many tasks in natural language processing\cite{worsham2020multi}, including named entity recognition. In multi-task learning, we generally choose several similar tasks with different models, and the hidden layers of these models are partly shared. In this way, models can provide better representations that are suitable for all tasks and avoid overfitting.

For multi-task learning in Arabic, Jarrar et al.\cite{jarrar2022wojood} convert the nested Arabic NER into a multi-task paradigm and present a multi-task model to predict entities on Wojood, a large-scale Arabic NER dataset collected from multiple sources including web articles, archives, and social media, covering both MSA and dialect. Specifically, they adopt AraBert\cite{antoun2020arabert} as the shared encoder and design different linear layers for each kind of entity type, which is used to predict corresponding BIO tags. In this way, the model can extract entities of different types independently. However, they still can not solve the nested entities from the same type. Eventually, this method can achieve an overall F1 of 88.40 on Wojood.

In addition, Ahmed et al.\cite{ahmed2022tafsir} combine the Arabic NER task with the topic modeling task. They accordingly present a dataset Tafsir as the first large-scale multi-task benchmark for Classical Arabic. These two tasks share the same embeddings. Then, they adopt a BiLSTM-CRF model for the upper layer NER task while using the linear layer for each label of the topic modeling task. 
Experiments show that the proposed method works effectively on both tasks.

% Multi-task learning is widely adopted in named entity recognition especially in some special scenarios. Wang et al.\cite{wang2019cross} take the basic BiLSTM-CRF model to extract entities in biomedical texts. They train the model on different datasets as an augmentation method while sharing char-level BiLSTM, word-level BiLSTM or the both parameters. Experiments demonstrate that sharing both word-level and char-level LSTM outperforms other baselines. Also, Li et al.\cite{crichton2017neural} propose the similar method. They use multiple annotated datasets and regard diverse datasets as different tasks. They propose two types of models. The first one share the basic hidden layers and predict with independent output layers for different tasks. The As for the second one, the output of the POS task is utilized as the input of the NER task. In most datasets, multi-task learning performs better than the single task model. Additionally, Zhou et al.\cite{zhou2021end} combine the named entity recognition task and the named entity normalization task in order to avoid error propagation. Two tasks are interconnected through shared features in medical domain. The results outperform other approaches as well. Futhermore, Aguilar et al.\cite{aguilar2017multi} incorporates multi-task learning for NER taks in social media domain. They design an named entity segmentation task as the auxiliary, which determines whether a given token is a named entity or not. And experiments show this task does contributes a lot.

\subsubsection{Transfer Learning for NER}

Low-resource languages often suffer from the scarcity of training data, which can be effectively alleviated by transfer learning\cite{houlsby2019parameter,zhuang2020comprehensive}. Especially, transfer learning on pre-trained models can benefit many NLP tasks due to the knowledge transferred across tasks or domains.

\begin{figure}[tp]
\centering
\includegraphics[width=0.48\textwidth]{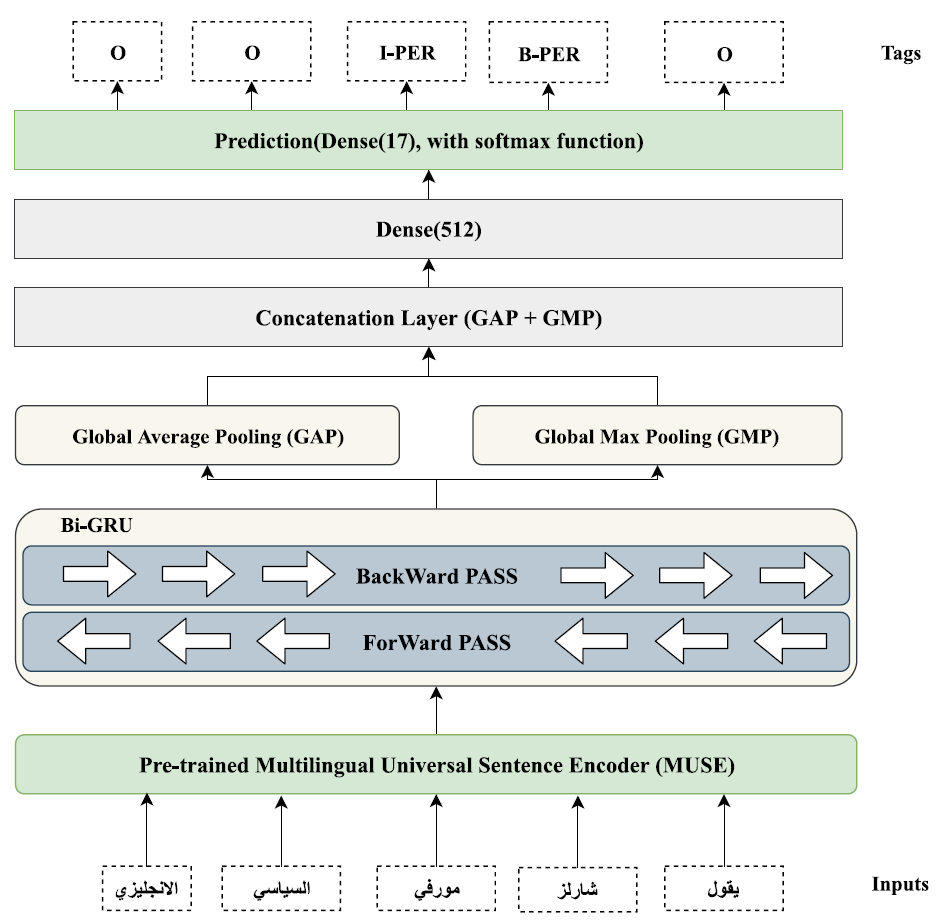}
\caption{\textbf{Architecture of Pooled-GRU Model\cite{al2020transfer}}.
}
\label{fig:transfer}
% \vspace{-4mm}
\end{figure}

For Arabic NER, to leverage the knowledge transfer ability from the multilingual pre-trained model, Al-Smadi et al.\cite{al2020transfer} propose a Pooled-GRU network based on Multilingual Universal Sentence Encoder \cite{yang2019multilingual} (MUSE) as shown in Figure~\ref{fig:transfer}. 
Specifically, they adopt a traditional sequence labeling model and a pre-trained multilingual universal sentence encoder to perform Arabic NER. In this way, the knowledge from the MUSE is well transferred to the Pooled-GPU. Benefiting from the transferred knowledge, the semantic relationships among 
sentences and phrases in other linguistic scenarios can be applied in Arabic as well. Experiments are conducted on an Arabic dataset WikiFANE\cite{alotaibi2014hybrid} and this model demonstrates significant improvement compared with the BiLSTM-CRF model.

\begin{figure*}[tb]
    \centering
    \includegraphics[width=1\textwidth]{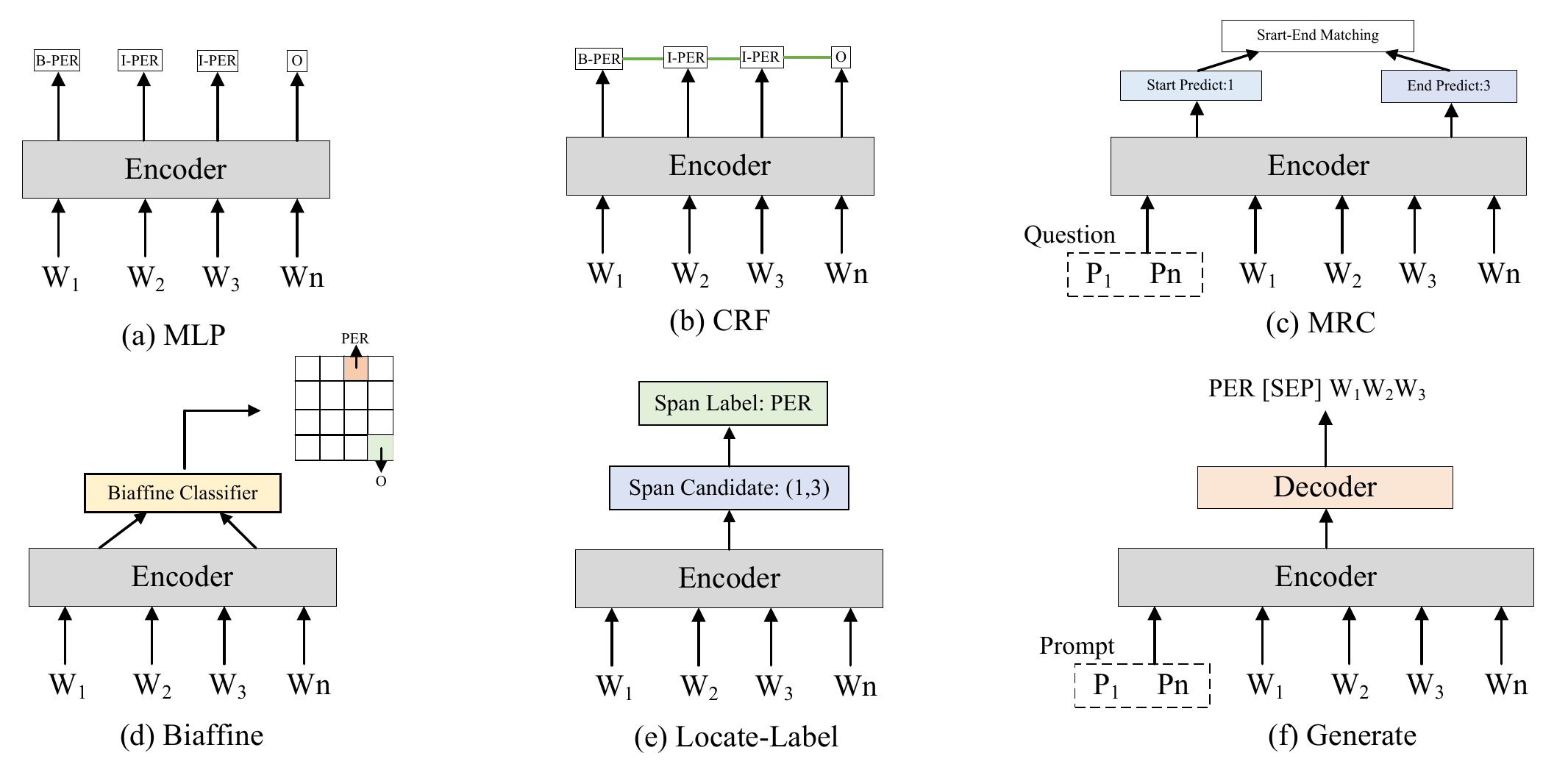}
    \caption{Architectures of different paradigms of NER solutions. (a) and (b) are adopted in previous Arabic NER methods. (c)-(f) are recent representative decoding strategies in English NER.}
    \label{fig:milesonte}
\end{figure*}

Besides the above multilingual MUSE, there are other multi-lingual transformers designed for cross-lingual tasks. Lan et al. \cite{lan2020empirical} present GigaBert, a bilingual Bert model, which is proposed for Arabic NLP tasks and English-to-Arabic transfer. This customized Bert is pre-trained on Gigaword and Wikipedia and some web crawl data. Specifically, the author customizes the vocabulary and creates the code-switched data to further strengthen the cross-lingual transfer ability. 
In this way, the model outperforms the popular multilingual Bert models in many NLP tasks including NER. In addition, with supervised Arabic data, GigaBert performs better than the state-of-art Arabic-specific Bert model.

% This paper propose a BGRU model based on BERT for Arabic named entity recognition\cite{alsaaran2021arabic}. The pre-trained model learns knowledge across domains with large unannotated data. The model is trained on ANERCorp\cite{benajiba2007anersys} and AQMAR\cite{mohit2012re} dataset. Since the two datasets use different labels to annotate person and miscellanies named entities, this paper unify the tags by preprocessing.

% \subsubsection{Deep Multi-Task Learning for NER
% }

\section{Future Direction}
With the development of modeling methods and the increase in downstream applications, we expect Arabic NER to achieve more progress in the future. Thus, we list the potential further directions for Arabic NER from the following five perspectives. 
1) The NER methods in English or Chinese have achieved significant progress in the past few years. The developments mainly focus on designing different label decoding strategies to adapt to diverse label structures. These important technology breakouts could be applied to Arabic NER but are not well explored. 
2) Besides innovative technologies, particular Arabic NER tasks need to be further investigated by the requirement of the downstream applications, for instance, the nested NER. The nested NER is well studied in English but only a few works of Arabic NER consider this task.   
3) Arabic NER, as a task with a small amount of available annotated data, also demands low-resource or distantly-supervised methods to relieve the problem of lacking data. In this way, Arabic NER can achieve a more robust performance under a few labeled annotated data.
4) Apart from leveraging more data to obtain a more robust Arabic NER model, cross-linguistic knowledge transfer also provides comprehensive information for model learning. To this end, we will analyze what cross-linguistic knowledge transfer technologies may benefit Arabic NER.
5) By jointly training NER with other auxiliary tasks, both two tasks can benefit from each other. Thus, we also discuss several tasks interrelated with Arabic NER. 
We will describe these topics in the following sections.

\textbf{Optional Label Decoding Strategies.} 
As described in Section 3.3, nearly all methods for Arabic NER adopt linear classification or CRF layer for label decoding. 
As shown in Figure 8(a) and Figure 8(b), the linear layer or CRF decodes the label for each token. 
Although these two decoding strategies achieved decent performance in flat NER, they fail to process more complex NER tasks, such as nested NER.  
Different from these two paradigms, actually, in English or Chinese NER, there are many different label decoding strategies for NER as depicted in Figure 8(c)-(f). These newly proposed frameworks can unify different forms of NER tasks.
Here we will simply introduce these representative solutions as follows: 

\begin{itemize}

\item \textbf{MRC}\footnote{https://github.com/ShannonAI/mrc-for-flat-nested-ner} As shown in Figure 8(c), Li et al. \cite{li2020unified} convert the NER task into a machine reading comprehension paradigm by inserting a question ahead of the sentence. The essence of using the MRC framework is to guide the model to search for the most relevant answers from the given text by designing NER label-related questions.
During decoding, they identify the corresponding start and end positions for the type of entity mentioned by the question. 

\item \textbf{Biaffine}\footnote{https://github.com/juntaoy/biaffine-ner} In Figure 8(d), Yu et al. \cite{yu2020named} recast the label decoding process into a dependency parsing task, where all pairs of start and end positions in the sentence are jointly considered and classified into specific types. The biaffine architecture is very effective and a lot of recent works have used it to further improve their performance \cite{li2022unified,gu-etal-2022-delving}.

\item \textbf{Locate-Label}\footnote{https://github.com/tricktreat/locate-and-label} As depicted in Figure 8(e), different from previous one-stage paradigms, Shen et al. \cite{shen2021locate} propose a two-stage solution that first locates the potential entities in the sentence and then identifies the specific class of these spans. 
This framework can effectively utilize the boundary information of entities. Xia et al. \cite{xia2019multi} also adopt similar methods to process nested NER.
% This framework is two-stage and may suffer from error propagation. 

% \item \textbf{Prompt} \cite{cui2021template, chen2021lightner} construct a template for each token in the sentence by asking the specific class of the token. Subsequently, they predict the corresponding content and project it into the entity type. This paradigm is to introduce external knowledge explicitly for the pre-trained language models, which is slow but effective for low-resource scenarios. 

\item \textbf{Generate}\footnote{https://github.com/universal-ie/UIE} In addition to the above encoder-based NER paradigms, recent encoder-decoder based methods attract increasing attention. As demonstrated in Figure 8(f), 
Lu et al. \cite{lu2022unified} transform the NER task into a generation task based on the generative pre-trained language model, such as T5 \cite{raffel2020exploring} or BART \cite{lewis2020bart}. In this way, the model does not need to predict the entity class for each token but directly generates the entity in the decoded output sentence. Besides generating entity words, there are also generative methods \cite{yan2021unified,chen2021lightner} which generate the indexes of entities in the original texts. Benefiting from the superior generative ability, these methods are unified for all kinds of NER tasks, including flat, nested, and discontinuous NER. It is worth noting that the performance of this paradigm may be heavily affected by user-constructed hard prompts. Therefore, how to design an appropriate prompt is also a challenge for Arabic NER.  

\end{itemize}

\textbf{Particular NER Tasks.} 
As described in the method section, most existing methods for Arabic NER only consider coarse-grained NER types in the general domain and focus on flat NER. 
Actually, in real applications, such as slot filling \cite{goo2018slot,chen2019bert,weld2021survey} in conversation, the entities are fine-grained and hierarchical, which means a named entity may belong to multiple types. 
Another challenge is that the entity could be nested, where a part of one entity corresponds to another entity type. 
In addition, the entities in the text may also be discontinuous, especially in the clinical corpus. Under this setting, researchers need to move beyond continuous entities and propose new methods to extract discontinuous ones.
Thus, we can resort to mainstream methods of English NER to solve these challenges in Arabic NER. Here we briefly describe several solutions for these challenges.

\begin{itemize}
    \item \textbf{Fine-grained NER.} To explore fine-grained NER where hundreds of named entity categories must be recognized, Mai et al. \cite{mai2018empirical} utilize the dictionary and category embeddings incorporating the neural network. Peng et al. \cite{peng2020toward} propose to adjust the existing
    NER system using the previously labeled data and entity lexicons of the newly introduced entity types. In this way, they can quickly adapt to fine-grained  entities.
    
    \item \textbf{Nested NER.} As shown in Table 4, although Arabic NER methods adopting CRF as a decoder achieves better performance than linear in flat NER datasets, CRF architecture is hard to work in nested NER. 
    To solve the nested NER, we can use the above proposed alternative solutions for NER, such as MRC, Biaffine, Locate-Label or Generate. There are also several graph-based methods that build an entity-entity graph \cite{wan2022nested} or model the nested ner as latent lexicalized constituency parsing \cite{lou2022nested}. With these methods, we can solve Arabic nested NER in a more comfortable way.
    
    \item \textbf{Discontinuous NER.} 
    % This kind of NER task  requires identifying all discontinuous entities in the given text. 
    Traditional NER is presented as a sequence-labeling task and hard to handle these entities. 
    Li et al. \cite{li2021span} propose a span-based model and  perform relation classification to determine whether a given pair of entity fragments are overlapping or succession. 
    Wang et al. \cite{wang2021discontinuous} build a segment graph for each sentence, in which each node denotes a segment and an edge connects two nodes that belong to the same entity. Thus, they can identify the discontinuous entities. There are also several transition-based method \cite{dai2020effective} or grid-tagging method \cite{liu2022toe} proposed for this task.
    Moreover, the generation-based model can directly output discontinuous entities.
     
\end{itemize}

\textbf{Low-resource and Distantly-supervised NER.} 
In fact, the open-source labeled dataset and available Arabic entity gazetteer of Arabic NER are very scarce. Labeling a large amount of NER data is both time-consuming and expensive. This predicament leads to unsatisfactory performance in practical applications of Arabic NER. To make the best use of limited labeled data, it is feasible to consider low-resource and distantly-supervised methods to achieve better results. 
Here we present several effective methods from English that may work for Arabic NER.

% \item \textbf{Prompt} \cite{cui2021template, chen2021lightner} construct a template for each token in the sentence by asking the specific class of the token. Subsequently, they predict the corresponding content and project it into the entity type. This paradigm is to introduce external knowledge explicitly for the pre-trained language models, which is slow but effective for low-resource scenarios. 

\begin{itemize}
    \item \textbf{Low-resource NER.} There are many works that utilize data augmentation to improve low-resource performance. 
    Dai et al. \cite{dai2020analysis} use label-wise token replacement, synonym replacement, and mention replacement to obtain new training samples. Ding et al. \cite{ding2020daga} model the linearized labeled sentences to generate new examples. 
    Liu et al. \cite{liu22low} present an augmentation approach that can elicit knowledge from BERT and does not need considerable human intervention. With the augmentation method, we can obtain much more training examples. Besides generating more training samples, there are also several recent methods \cite{cui2021template,lee2022good} focusing on designing low-resource algorithms. They effectively leverage the knowledge from the large pre-trained language models and achieve significant results.   

    \item \textbf{Distantly-supervised NER.} With the help of Arabic gazetteer or existing NER tools, we can easily obtain abundant weak-labeled samples by matching or labeling unlabeled data. However, such weak labels are noisy. To comprehensively utilize these noisy samples, Yang et al. \cite{yang2018distantly} propose a partial annotation learning paradigm. More recently, \cite{peng2019distantly} introduces PU-learning to label positive tokens instead of directly learning the weak labels. Teacher-student frameworks \cite{liang2020bond, meng2021distantly,zhang2021improving,qu2022distantly} are widely adopted to refine noisy labels and achieve superior performance. To jointly learn from a small amount of labeled data and large weakly labeled data, Jiang et al. \cite{jiang2021named} propose a multi-stage computational framework to achieve robust learning.
    
\end{itemize}

\textbf{Cross-linguistic Knowledge Transfer.} Transferring knowledge from high-resource language to low-resource language is proven effective to alleviate the data scarcity problem. To enhance the capabilities of the Arabic NER system, it is worth considering knowledge transfer from other languages. To perform cross-linguistic knowledge transfer, here we present several effective solutions. 

\begin{itemize}
    \item \textbf{Machine Translation.} Currently, over 100 languages can be supported by strong machine translation systems. Machine translation is a straightforward way to translate training sentences and entities from other languages into target languages.
    % Currently, over 100 languages can be supported by strong machine translation systems. However, only a subset of them possesses large annotated corpora for NER. 
    Thus, machine translation can be leveraged to improve performance to cross-lingual NER \cite{jain2019entity,dai2020analysis,feng2021survey}. 
    
    \item \textbf{Adversarial Training.} A language-adversarial task during finetuning can significantly improve the cross-lingual transfer performance \cite{keung2019adversarial}. In other words, the adversarial techniques encourage the model to align the representations of English documents and their translations of target languages. Furthermore, the adversarial learning framework can leverage the unlabeled data and improve the result \cite{chen2021advpicker}, where a discriminator picks less language-dependent target language data according to the similarity to the source language.

    \item \textbf{Multilingual Pretrained Language Model.} The multilingual pre-trained language models, such as mBERT, XLM \cite{lample2019cross}, mT5 \cite{xue2021mt5},
    provide a straightforward way to enable zero-shot learning via cross-lingual transfer, thus eliminating the need for labeled data for the target task and language \cite{ebrahimi2021adapt}. For pre-trained languages, the cross-lingual transfer is especially efficient \cite{pires2019multilingual}.

\end{itemize}

\textbf{Jointly training NER with other tasks.} 
There are many tasks that are semantically related to NER. It is worth exploring approaches for jointly performing NER and other tasks. Here we introduce most related tasks. 

\begin{itemize}
    \item \textbf{NER and Entity Linking}. Most entity linking systems discard mention detection, performing only entity disambiguation of previously detected mentions. Thus, the dependency between the entity linking and NER is ignored. In real scenarios, we can do joint learning of NER and
    EL in order to leverage the information of both tasks at every decision \cite{sil2013re, luo2015joint, martins2019joint}. Furthermore, by having a flow of information between the computation of the representations used for NER and entity linking, we are able to improve the model.

    \item \textbf{NER and Relation Extraction}. Both NER and relation extraction aim to extract structured information
    from unstructured texts. The typical pipeline approach \cite{zelenko2003kernel,chan2011exploiting} is to first identify entity mentions, and next perform
    classification between every two mentions to extract relations. However, this pipeline leads to the error
    propagation issue. To this end, an alternative and
    more recent approach is to perform these two tasks
    jointly \cite{miwa2016end, wang2020two}. Thus, we can leverage the interaction between tasks, resulting in improved performance. 

\end{itemize}

\textbf{Dialect Arabic NER.} {In the method section, most methods focus on modern standard Arabic (MSA). However, there are a large number of dialects in Arabic texts. Despite some similar
characteristics between MSA and Dialectal Arabic (DA), the significant differences between the two language varieties hinder
MSA systems from solving NER for Dialectal Arabic. Thus, 
it is a necessary and promising direction to improve the performance of dialect Arabic NER. Here we present a few effective methods for dialect Arabic NER.}

\begin{itemize}

\item \textbf{Pretrained Language Model for Dialects.} There are several pretrained Arabic PLM that utilize dialects for training. Their training corpus usually contains Arabic tweets consisting of multiple dialects. Specifically, MARBERT \footnote{https://github.com/UBC-NLP/marbert} randomly sample 1B
Arabic tweets from a large in-house dataset of
about 6B tweets. AraBERTv0.2-Twitter \footnote{https://github.com/aub-mind/arabert} trains on 60M Arabic tweets filtered from a collection of 100M. There is also 
dedicated pretrained model DziriBERT for the Algerian Dialect \cite{abdaoui2021dziribert}. It is feasible to continue pretraining on specific dialects for downstream tasks. 

\item \textbf{Unsupervised Domain Adaptation for Dialects.} Unsupervised Domain Adaptation (UDA) can be a first-order candidate approach for cross-dialect learning. Mekki et al. \cite{el2022adasl} introduce an unsupervised domain adaptation framework AdaSL for Arabic multi-dialectal sequence labeling, leveraging unlabeled DA data and labeled MSA data. They achieve new state-of-the-art zero-shot performance for Arabic multi-dialectal sequence labeling. Thus, it is a direction worth exploring to leverage MSA for Arabic dialects by domain adaptation.

\end{itemize}

% one effective solution is back-translation, which translate texts from other language to Arabic.  

% \textbf{Easy-to-Use Toolkit for Arabic NER.}

\section{Conclusion}
As the development of Arabic named entity recognition is significantly below the English counterpart, this survey aims to provide a comprehensive review for Arabic named entity recognition to help researchers understand this field. Specifically, we introduce the background of Arabic NER, especially the unique challenges of Arabic NER compared to English NER. Then, we summarize the current resources for Arabic NER, including the datasets and popular processing tools. With these resources, researchers can start their exploration easily and have a quick glance at Arabic NLP. To help straighten out the development of Arabic NER methods, we carefully categorize the Arabic NER methods into four paradigms, i.e., rule-based methods, machine learning methods, deep learning methods, and pre-trained language model methods. In these four paradigms, we present the most representative methods and give a description of their architectures. Furthermore, we summarize the applications of Arabic NER which promote the improvements of downstream tasks. Finally, we highlight several potential directions for Arabic NER. We hope this survey can provide handy guidance when designing Arabic NER models and attract more researchers to contribute to this field.

% if have a single appendix:
%\appendix[Proof of the Zonklar Equations]
% or
%\appendix  % for no appendix heading
% do not use \section anymore after \appendix, only \section*
% is possibly needed

% use appendices with more than one appendix
% then use \section to start each appendix
% you must declare a \section before using any
% \subsection or using \label (\appendices by itself
% starts a section numbered zero.)
%

% \appendices
% \section{Proof of the First Zonklar Equation}
% Appendix one text goes here.

% you can choose not to have a title for an appendix
% if you want by leaving the argument blank
% \section{}
% Appendix two text goes here.

% use section* for acknowledgment
% \ifCLASSOPTIONcompsoc
%   % The Computer Society usually uses the plural form
%   \section*{Acknowledgments}
% \else
%   % regular IEEE prefers the singular form
%   \section*{Acknowledgment}
% \fi

% The authors would like to thank...

% Can use something like this to put references on a page
% by themselves when using endfloat and the captionsoff option.
\ifCLASSOPTIONcaptionsoff
  \newpage
\fi

% trigger a \newpage just before the given reference
% number - used to balance the columns on the last page
% adjust value as needed - may need to be readjusted if
% the document is modified later
%\IEEEtriggeratref{8}
% The "triggered" command can be changed if desired:
%\IEEEtriggercmd{\enlargethispage{-5in}}

% references section

% can use a bibliography generated by BibTeX as a .bbl file
% BibTeX documentation can be easily obtained at:
% http://mirror.ctan.org/biblio/bibtex/contrib/doc/
% The IEEEtran BibTeX style support page is at:
% http://www.michaelshell.org/tex/ieeetran/bibtex/
\bibliographystyle{IEEEtran}
% argument is your BibTeX string definitions and bibliography database(s)
\bibliography{reference}

% Generated by IEEEtran.bst, version: 1.14 (2015/08/26)
\begin{thebibliography}{100}
\providecommand{\url}[1]{#1}
\csname url@samestyle\endcsname
\providecommand{\newblock}{\relax}
\providecommand{\bibinfo}[2]{#2}
\providecommand{\BIBentrySTDinterwordspacing}{\spaceskip=0pt\relax}
\providecommand{\BIBentryALTinterwordstretchfactor}{4}
\providecommand{\BIBentryALTinterwordspacing}{\spaceskip=\fontdimen2\font plus
\BIBentryALTinterwordstretchfactor\fontdimen3\font minus
  \fontdimen4\font\relax}
\providecommand{\BIBforeignlanguage}[2]{{%
\expandafter\ifx\csname l@#1\endcsname\relax
\typeout{** WARNING: IEEEtran.bst: No hyphenation pattern has been}%
\typeout{** loaded for the language `#1'. Using the pattern for}%
\typeout{** the default language instead.}%
\else
\language=\csname l@#1\endcsname
\fi
#2}}
\providecommand{\BIBdecl}{\relax}
\BIBdecl

\bibitem{guellil2021arabic}
I.~Guellil, H.~Sa{\^a}dane, F.~Azouaou, B.~Gueni, and D.~Nouvel, ``Arabic
  natural language processing: An overview,'' \emph{Journal of King Saud
  University-Computer and Information Sciences}, vol.~33, no.~5, pp. 497--507,
  2021.

\bibitem{cheng2021hacred}
Q.~Cheng, J.~Liu, X.~Qu, J.~Zhao, J.~Liang, Z.~Wang, B.~Huai, N.~J. Yuan, and
  Y.~Xiao, ``Hacred: A large-scale relation extraction dataset toward hard
  cases in practical applications,'' in \emph{Findings of the Association for
  Computational Linguistics: ACL-IJCNLP 2021}, 2021, pp. 2819--2831.

\bibitem{gu2021read}
Y.~Gu, X.~Qu, Z.~Wang, B.~Huai, N.~J. Yuan, and X.~Gui, ``Read, retrospect,
  select: An mrc framework to short text entity linking,'' in \emph{Proceedings
  of the AAAI Conference on Artificial Intelligence}, vol.~35, no.~14, 2021,
  pp. 12\,920--12\,928.

\bibitem{zhu2021efficient}
T.~Zhu, X.~Qu, W.~Chen, Z.~Wang, B.~Huai, N.~J. Yuan, and M.~Zhang, ``Efficient
  document-level event extraction via pseudo-trigger-aware pruned complete
  graph,'' \emph{arXiv preprint arXiv:2112.06013}, 2021.

\bibitem{clark2016improving}
K.~Clark and C.~D. Manning, ``Improving coreference resolution by learning
  entity-level distributed representations,'' in \emph{Proceedings of the 54th
  Annual Meeting of the Association for Computational Linguistics (Volume 1:
  Long Papers)}, 2016, pp. 643--653.

\bibitem{ugawa2018neural}
A.~Ugawa, A.~Tamura, T.~Ninomiya, H.~Takamura, and M.~Okumura, ``Neural machine
  translation incorporating named entity,'' in \emph{Proceedings of the 27th
  International Conference on Computational Linguistics}, 2018, pp. 3240--3250.

\bibitem{benajiba2008arabica}
Y.~Benajiba and P.~Rosso, ``Arabic named entity recognition using conditional
  random fields,'' in \emph{Proc. of Workshop on HLT \& NLP within the Arabic
  World, LREC}, vol.~8, 2008, pp. 143--153.

\bibitem{elgibali2005investigating}
A.~Elgibali, \emph{Investigating Arabic: Current parameters in analysis and
  learning}.\hskip 1em plus 0.5em minus 0.4em\relax Brill, 2005, vol.~42.

\bibitem{shaalan2009nera}
K.~Shaalan and H.~Raza, ``Nera: Named entity recognition for arabic,''
  \emph{Journal of the American Society for Information Science and
  Technology}, vol.~60, no.~8, pp. 1652--1663, 2009.

\bibitem{shaalan2014survey}
K.~Shaalan, ``A survey of arabic named entity recognition and classification,''
  \emph{Computational Linguistics}, vol.~40, no.~2, pp. 469--510, 2014.

\bibitem{zirikly2015named}
A.~Zirikly and M.~Diab, ``Named entity recognition for arabic social media,''
  in \emph{Proceedings of the 1st workshop on vector space modeling for natural
  language processing}, 2015, pp. 176--185.

\bibitem{dandashi2016arabic}
A.~Dandashi, J.~A. Jaam, and S.~Foufou, ``Arabic named entity recognition—a
  survey and analysis,'' in \emph{Intelligent Interactive Multimedia Systems
  and Services 2016}.\hskip 1em plus 0.5em minus 0.4em\relax Springer, 2016,
  pp. 83--96.

\bibitem{salah2017comparative}
R.~E. Salah and L.~Q. binti Zakaria, ``A comparative review of machine learning
  for arabic named entity recognition,'' \emph{International Journal on
  Advanced Science, Engineering and Information Technology}, vol.~7, no.~2, pp.
  511--518, 2017.

\bibitem{el2019arabic}
I.~El~Bazi and N.~Laachfoubi, ``Arabic named entity recognition using deep
  learning approach.'' \emph{International Journal of Electrical \& Computer
  Engineering (2088-8708)}, vol.~9, no.~3, 2019.

\bibitem{liu2019arabic}
L.~Liu, J.~Shang, and J.~Han, ``Arabic named entity recognition: What works and
  what’s next,'' in \emph{Proceedings of the Fourth Arabic Natural Language
  Processing Workshop}, 2019, pp. 60--67.

\bibitem{ali2020recent}
B.~A.~B. Ali, S.~Mihi, I.~El~Bazi, and N.~Laachfoubi, ``A recent survey of
  arabic named entity recognition on social media.'' \emph{Rev. d'Intelligence
  Artif.}, vol.~34, no.~2, pp. 125--135, 2020.

\bibitem{li2020survey}
J.~Li, A.~Sun, J.~Han, and C.~Li, ``A survey on deep learning for named entity
  recognition,'' \emph{IEEE Transactions on Knowledge and Data Engineering},
  vol.~34, no.~1, pp. 50--70, 2020.

\bibitem{yadav2018survey}
V.~Yadav and S.~Bethard, ``A survey on recent advances in named entity
  recognition from deep learning models,'' in \emph{Proceedings of the 27th
  International Conference on Computational Linguistics}, 2018, pp. 2145--2158.

\bibitem{algahtani2012arabic}
S.~Algahtani, \emph{Arabic named entity recognition: a corpus-based
  study}.\hskip 1em plus 0.5em minus 0.4em\relax The University of Manchester
  (United Kingdom), 2012.

\bibitem{abdelrahman2010integrated}
S.~AbdelRahman, M.~Elarnaoty, M.~Magdy, and A.~Fahmy, ``Integrated machine
  learning techniques for arabic named entity recognition,'' \emph{IJCSI},
  vol.~7, no.~4, pp. 27--36, 2010.

\bibitem{benajiba2007anersys2}
Y.~Benajiba and P.~Rosso, ``Anersys 2.0: Conquering the ner task for the arabic
  language by combining the maximum entropy with pos-tag information.'' in
  \emph{IICAI}, 2007, pp. 1814--1823.

\bibitem{shaalan2007person}
K.~Shaalan and H.~Raza, ``Person name entity recognition for arabic,'' in
  \emph{Proceedings of the 2007 workshop on computational approaches to semitic
  languages: common issues and resources}, 2007, pp. 17--24.

\bibitem{steinberger2012survey}
R.~Steinberger, ``A survey of methods to ease the development of highly
  multilingual text mining applications,'' \emph{Language resources and
  evaluation}, vol.~46, no.~2, pp. 155--176, 2012.

\bibitem{alkharashi2009person}
Alkharashi and Ibrahim, ``Person named entity generation and recognition for
  arabic language,'' in \emph{Proceedings of the Second International
  Conference on Arabic Language Resources and Tools}, 2009, pp. 205--208.

\bibitem{benajiba2009arabic}
Y.~Benajiba, M.~Diab, and P.~Rosso, ``Arabic named entity recognition: A
  feature-driven study,'' \emph{IEEE Transactions on Audio, Speech, and
  Language Processing}, vol.~17, no.~5, pp. 926--934, 2009.

\bibitem{antoun2020arabert}
W.~Antoun, F.~Baly, and H.~Hajj, ``Arabert: Transformer-based model for arabic
  language understanding,'' in \emph{Proceedings of the 4th Workshop on
  Open-Source Arabic Corpora and Processing Tools, with a Shared Task on
  Offensive Language Detection}, 2020, pp. 9--15.

\bibitem{antoun2021araelectra191}
------, ``Araelectra: Pre-training text discriminators for arabic language
  understanding,'' in \emph{Proceedings of the Sixth Arabic Natural Language
  Processing Workshop}, 2021, pp. 191--195.

\bibitem{oudah2012pipeline}
M.~Oudah and K.~Shaalan, ``A pipeline arabic named entity recognition using a
  hybrid approach,'' in \emph{Proceedings of COLING 2012}, 2012, pp.
  2159--2176.

\bibitem{benajiba2007anersys}
Y.~Benajiba, P.~Rosso, and J.~M. Bened{\'\i}ruiz, ``Anersys: An arabic named
  entity recognition system based on maximum entropy,'' in \emph{International
  Conference on Intelligent Text Processing and Computational
  Linguistics}.\hskip 1em plus 0.5em minus 0.4em\relax Springer, 2007, pp.
  143--153.

\bibitem{abdul2010simplified}
A.~Abdul-Hamid and K.~Darwish, ``Simplified feature set for arabic named entity
  recognition,'' in \emph{Proceedings of the 2010 named entities workshop},
  2010, pp. 110--115.

\bibitem{zaghouani2012renar}
W.~Zaghouani, ``Renar: A rule-based arabic named entity recognition system,''
  \emph{ACM Transactions on Asian Language Information Processing (TALIP)},
  vol.~11, no.~1, pp. 1--13, 2012.

\bibitem{morsi2013studying}
A.~Morsi and A.~Rafea, ``Studying the impact of various features on the
  performance of conditional random field-based arabic named entity
  recognition,'' in \emph{2013 ACS International Conference on Computer Systems
  and Applications (AICCSA)}.\hskip 1em plus 0.5em minus 0.4em\relax IEEE,
  2013, pp. 1--5.

\bibitem{mohit2012recall}
B.~Mohit, N.~Schneider, R.~Bhowmick, K.~Oflazer, and N.~A. Smith,
  ``Recall-oriented learning of named entities in arabic wikipedia,'' in
  \emph{Proceedings of the 13th Conference of the European Chapter of the
  Association for Computational Linguistics}, 2012, pp. 162--173.

\bibitem{salah2018building}
R.~E. Salah and L.~Q.~B. Zakaria, ``Building the classical arabic named entity
  recognition corpus (canercorpus),'' in \emph{2018 Fourth International
  Conference on Information Retrieval and Knowledge Management (CAMP)}.\hskip
  1em plus 0.5em minus 0.4em\relax IEEE, 2018, pp. 1--8.

\bibitem{pasha2014madamira}
A.~Pasha, M.~Al-Badrashiny, M.~Diab, A.~El~Kholy, R.~Eskander, N.~Habash,
  M.~Pooleery, O.~Rambow, and R.~Roth, ``Madamira: A fast, comprehensive tool
  for morphological analysis and disambiguation of arabic,'' in
  \emph{Proceedings of the ninth international conference on language resources
  and evaluation (LREC'14)}, 2014, pp. 1094--1101.

\bibitem{monroe-etal-2014-word}
\BIBentryALTinterwordspacing
W.~Monroe, S.~Green, and C.~D. Manning, ``Word segmentation of informal
  {A}rabic with domain adaptation,'' in \emph{Proceedings of the 52nd Annual
  Meeting of the Association for Computational Linguistics (Volume 2: Short
  Papers)}.\hskip 1em plus 0.5em minus 0.4em\relax Baltimore, Maryland:
  Association for Computational Linguistics, Jun. 2014, pp. 206--211. [Online].
  Available: \url{https://aclanthology.org/P14-2034}
\BIBentrySTDinterwordspacing

\bibitem{abdelali-etal-2016-farasa}
\BIBentryALTinterwordspacing
A.~Abdelali, K.~Darwish, N.~Durrani, and H.~Mubarak, ``{F}arasa: A fast and
  furious segmenter for {A}rabic,'' in \emph{Proceedings of the 2016 Conference
  of the North {A}merican Chapter of the Association for Computational
  Linguistics: Demonstrations}.\hskip 1em plus 0.5em minus 0.4em\relax San
  Diego, California: Association for Computational Linguistics, Jun. 2016, pp.
  11--16. [Online]. Available: \url{https://aclanthology.org/N16-3003}
\BIBentrySTDinterwordspacing

\bibitem{obeid2020camel}
O.~Obeid, N.~Zalmout, S.~Khalifa, D.~Taji, M.~Oudah, B.~Alhafni, G.~Inoue,
  F.~Eryani, A.~Erdmann, and N.~Habash, ``Camel tools: An open source python
  toolkit for arabic natural language processing,'' in \emph{Proceedings of the
  12th language resources and evaluation conference}, 2020, pp. 7022--7032.

\bibitem{grishman-sundheim-1996-message}
\BIBentryALTinterwordspacing
R.~Grishman and B.~Sundheim, ``{M}essage {U}nderstanding {C}onference- 6: A
  brief history,'' in \emph{{COLING} 1996 Volume 1: The 16th International
  Conference on Computational Linguistics}, 1996. [Online]. Available:
  \url{https://aclanthology.org/C96-1079}
\BIBentrySTDinterwordspacing

\bibitem{abuleil2004extracting}
S.~Abuleil and M.~Evens, ``Extracting names from arabic text for
  question-answering systems.'' in \emph{RIAO}, 2004, pp. 638--647.

\bibitem{mesfar2007named}
S.~Mesfar, ``Named entity recognition for arabic using syntactic grammars,'' in
  \emph{Natural Language Processing and Information Systems: 12th International
  Conference on Applications of Natural Language to Information Systems, NLDB
  2007, Paris, France, June 27-29, 2007. Proceedings 12}.\hskip 1em plus 0.5em
  minus 0.4em\relax Springer, 2007, pp. 305--316.

\bibitem{shaalan2008arabic}
K.~Shaalan and H.~Raza, ``Arabic named entity recognition from diverse text
  types,'' in \emph{International conference on natural language
  processing}.\hskip 1em plus 0.5em minus 0.4em\relax Springer, 2008, pp.
  440--451.

\bibitem{elsebai2011extracting}
A.~Elsebai and F.~Meziane, ``Extracting person names from arabic newspapers,''
  in \emph{2011 International Conference on Innovations in Information
  Technology}.\hskip 1em plus 0.5em minus 0.4em\relax IEEE, 2011, pp. 87--89.

\bibitem{abdallah2012integrating}
S.~Abdallah, K.~Shaalan, and M.~Shoaib, ``Integrating rule-based system with
  classification for arabic named entity recognition,'' in \emph{International
  Conference on Intelligent Text Processing and Computational
  Linguistics}.\hskip 1em plus 0.5em minus 0.4em\relax Springer, 2012, pp.
  311--322.

\bibitem{maloney1998tagarab}
J.~Maloney and M.~Niv, ``Tagarab: a fast, accurate arabic name recognizer using
  high-precision morphological analysis,'' in \emph{Computational approaches to
  semitic languages}, 1998.

\bibitem{traboulsi2009arabic}
H.~Traboulsi, ``Arabic named entity extraction: A local grammar-based
  approach,'' in \emph{2009 International Multiconference on Computer Science
  and Information Technology}.\hskip 1em plus 0.5em minus 0.4em\relax IEEE,
  2009, pp. 139--143.

\bibitem{elsebairule}
A.~Elsebai, F.~Meziane, F.~Z. Belkredim, and U.~H.~B. Bouali, ``A rule based
  persons names arabic extraction system.''

\bibitem{buckwalter2004issues}
T.~Buckwalter, ``Issues in arabic orthography and morphology analysis,'' in
  \emph{proceedings of the workshop on computational approaches to Arabic
  script-based languages}, 2004, pp. 31--34.

\bibitem{harrag2011extracting}
F.~Harrag, E.~El-Qawasmeh, and A.~M. Salman Al-Salman, ``Extracting named
  entities from prophetic narration texts (hadith),'' in \emph{International
  Conference on Software Engineering and Computer Systems}.\hskip 1em plus
  0.5em minus 0.4em\relax Springer, 2011, pp. 289--297.

\bibitem{benajiba2008arabicb}
Y.~Benajiba, M.~Diab, P.~Rosso \emph{et~al.}, ``Arabic named entity
  recognition: An svm-based approach,'' in \emph{Proceedings of 2008 Arab
  international conference on information technology (ACIT)}.\hskip 1em plus
  0.5em minus 0.4em\relax Association of Arab Universities Amman, Jordan, 2008,
  pp. 16--18.

\bibitem{bidhendi2012extracting}
M.~A. Bidhendi, B.~Minaei-Bidgoli, and H.~Jouzi, ``Extracting person names from
  ancient islamic arabic texts,'' in \emph{Proceedings of Language Resources
  and Evaluation for Religious Texts (LRE-Rel) Workshop Programme, Eight
  International Conference on Language Resources and Evaluation (LREC 2012)},
  2012, pp. 1--6.

\bibitem{hastie2009elements}
T.~Hastie, R.~Tibshirani, J.~H. Friedman, and J.~H. Friedman, \emph{The
  elements of statistical learning: data mining, inference, and
  prediction}.\hskip 1em plus 0.5em minus 0.4em\relax Springer, 2009, vol.~2.

\bibitem{koulali2012contribution}
R.~Koulali and A.~Meziane, ``A contribution to arabic named entity
  recognition,'' in \emph{2012 Tenth International Conference on ICT and
  Knowledge Engineering}.\hskip 1em plus 0.5em minus 0.4em\relax IEEE, 2012,
  pp. 46--52.

\bibitem{lotfyenhancing}
A.~Lotfy, C.~Sabty, and S.~Abdennadher, ``Enhancing deep learning with embedded
  features for arabic named entity recognition.''

\bibitem{gridach2016deep}
M.~Gridach, ``Deep learning approach for arabic named entity recognition,'' in
  \emph{International Conference on Intelligent Text Processing and
  Computational Linguistics}.\hskip 1em plus 0.5em minus 0.4em\relax Springer,
  2016, pp. 439--451.

\bibitem{mikolov2013distributed}
T.~Mikolov, I.~Sutskever, K.~Chen, G.~S. Corrado, and J.~Dean, ``Distributed
  representations of words and phrases and their compositionality,''
  \emph{Advances in neural information processing systems}, vol.~26, 2013.

\bibitem{zhao2019recurrent}
Y.~Zhao, Y.~Shen, and J.~Yao, ``Recurrent neural network for text
  classification with hierarchical multiscale dense connections.'' in
  \emph{IJCAI}, 2019, pp. 5450--5456.

\bibitem{lin2021asrnn}
J.~C.-W. Lin, Y.~Shao, Y.~Djenouri, and U.~Yun, ``Asrnn: A recurrent neural
  network with an attention model for sequence labeling,''
  \emph{Knowledge-Based Systems}, vol. 212, p. 106548, 2021.

\bibitem{fi10120123}
\BIBentryALTinterwordspacing
M.~N.~A. Ali, G.~Tan, and A.~Hussain, ``Bidirectional recurrent neural network
  approach for arabic named entity recognition,'' \emph{Future Internet},
  vol.~10, no.~12, 2018. [Online]. Available:
  \url{https://www.mdpi.com/1999-5903/10/12/123}
\BIBentrySTDinterwordspacing

\bibitem{huang2015bidirectional}
Z.~Huang, W.~Xu, and K.~Yu, ``Bidirectional lstm-crf models for sequence
  tagging,'' \emph{arXiv preprint arXiv:1508.01991}, 2015.

\bibitem{shahina2019sequential}
K.~Shahina, P.~Jyothsna, G.~Prabha, B.~Premjith, and K.~Soman, ``A sequential
  labelling approach for the named entity recognition in arabic language using
  deep learning algorithms,'' in \emph{2019 International Conference on Data
  Science and Communication (IconDSC)}.\hskip 1em plus 0.5em minus 0.4em\relax
  IEEE, 2019, pp. 1--6.

\bibitem{gridach-2016-character}
M.~Gridach, ``Character-aware neural networks for {A}rabic named entity
  recognition for social media,'' in \emph{Proceedings of the 6th Workshop on
  South and Southeast {A}sian Natural Language Processing ({WSSANLP}2016)},
  Dec. 2016, pp. 23--32.

\bibitem{darwish2013named}
K.~Darwish, ``Named entity recognition using cross-lingual resources: Arabic as
  an example,'' in \emph{Proceedings of the 51st Annual Meeting of the
  Association for Computational Linguistics (Volume 1: Long Papers)}, 2013, pp.
  1558--1567.

\bibitem{awad2018arabic}
D.~Awad, C.~Sabty, M.~Elmahdy, and S.~Abdennadher, ``Arabic name entity
  recognition using deep learning,'' in \emph{International Conference on
  Statistical Language and Speech Processing}.\hskip 1em plus 0.5em minus
  0.4em\relax Springer, 2018, pp. 105--116.

\bibitem{ali2019boosting}
M.~N.~A. Ali, G.~Tan, and A.~Hussain, ``Boosting arabic named-entity
  recognition with multi-attention layer,'' \emph{IEEE Access}, vol.~7, pp.
  46\,575--46\,582, 2019.

\bibitem{helwe2019arabic}
C.~Helwe and S.~Elbassuoni, ``Arabic named entity recognition via deep
  co-learning,'' \emph{Artificial Intelligence Review}, vol.~52, no.~1, pp.
  197--215, 2019.

\bibitem{vaswani2017attention}
A.~Vaswani, N.~Shazeer, N.~Parmar, J.~Uszkoreit, L.~Jones, A.~N. Gomez,
  {\L}.~Kaiser, and I.~Polosukhin, ``Attention is all you need,''
  \emph{Advances in neural information processing systems}, vol.~30, 2017.

\bibitem{liu-etal-2019-arabic}
L.~Liu, J.~Shang, and J.~Han, ``{A}rabic named entity recognition: What works
  and what{'}s next,'' in \emph{Proceedings of the Fourth Arabic Natural
  Language Processing Workshop}.\hskip 1em plus 0.5em minus 0.4em\relax
  Florence, Italy: Association for Computational Linguistics, Aug. 2019, pp.
  60--67.

\bibitem{kenton2019bert}
J.~D. M.-W.~C. Kenton and L.~K. Toutanova, ``Bert: Pre-training of deep
  bidirectional transformers for language understanding,'' in \emph{Proceedings
  of NAACL-HLT}, 2019, pp. 4171--4186.

\bibitem{Clark2020ELECTRAPT}
K.~Clark, M.-T. Luong, Q.~V. Le, and C.~D. Manning, ``Electra: Pre-training
  text encoders as discriminators rather than generators,'' \emph{ArXiv}, vol.
  abs/2003.10555, 2020.

\bibitem{lan2019albert}
Z.~Lan, M.~Chen, S.~Goodman, K.~Gimpel, P.~Sharma, and R.~Soricut, ``Albert: A
  lite bert for self-supervised learning of language representations,''
  \emph{arXiv preprint arXiv:1909.11942}, 2019.

\bibitem{ali_safaya_2020_4718724}
\BIBentryALTinterwordspacing
A.~Safaya, ``Arabic-albert,'' Aug. 2020. [Online]. Available:
  \url{https://doi.org/10.5281/zenodo.4718724}
\BIBentrySTDinterwordspacing

\bibitem{ghaddar2022revisiting}
A.~Ghaddar, Y.~Wu, S.~Bagga, A.~Rashid, K.~Bibi, M.~Rezagholizadeh, C.~Xing,
  Y.~Wang, D.~Xinyu, Z.~Wang \emph{et~al.}, ``Revisiting pre-trained language
  models and their evaluation for arabic natural language understanding,''
  \emph{arXiv preprint arXiv:2205.10687}, 2022.

\bibitem{Antoun2021AraGPT2PT}
W.~Antoun, F.~Baly, and H.~M. Hajj, ``Aragpt2: Pre-trained transformer for
  arabic language generation,'' in \emph{WANLP}, 2021.

\bibitem{nagoudi-etal-2022-arat5}
E.~M.~B. Nagoudi, A.~Elmadany, and M.~Abdul-Mageed, ``{A}ra{T}5: Text-to-text
  transformers for {A}rabic language generation,'' in \emph{Proceedings of the
  60th Annual Meeting of the Association for Computational Linguistics (Volume
  1: Long Papers)}, May 2022, pp. 628--647.

\bibitem{Alsaaran2021ClassicalAN}
N.~Alsaaran and M.~AlRabiah, ``Classical arabic named entity recognition using
  variant deep neural network architectures and bert,'' \emph{IEEE Access},
  vol.~9, pp. 91\,537--91\,547, 2021.

\bibitem{al-qurishi-souissi-2021-arabic}
\BIBentryALTinterwordspacing
M.~S. Al-Qurishi and R.~Souissi, ``{A}rabic named entity recognition using
  transformer-based-{CRF} model,'' in \emph{Proceedings of The Fourth
  International Conference on Natural Language and Speech Processing (ICNLSP
  2021)}.\hskip 1em plus 0.5em minus 0.4em\relax Trento, Italy: Association for
  Computational Linguistics, 12--13 Nov. 2021, pp. 262--271. [Online].
  Available: \url{https://aclanthology.org/2021.icnlsp-1.31}
\BIBentrySTDinterwordspacing

\bibitem{helwe-etal-2020-semi}
C.~Helwe, G.~Dib, M.~Shamas, and S.~Elbassuoni, ``A semi-supervised {BERT}
  approach for {A}rabic named entity recognition,'' in \emph{Proceedings of the
  Fifth Arabic Natural Language Processing Workshop}, 2020, pp. 49--57.

\bibitem{el2013kalimat}
M.~El-Haj and R.~Koulali, ``Kalimat a multipurpose arabic corpus,'' in
  \emph{Second workshop on Arabic corpus linguistics (WACL-2)}, 2013, pp.
  22--25.

\bibitem{abdul2021arbert}
M.~Abdul-Mageed, A.~Elmadany, and E.~M.~B. Nagoudi, ``Arbert \& marbert: Deep
  bidirectional transformers for arabic,'' in \emph{Proceedings of the 59th
  Annual Meeting of the Association for Computational Linguistics and the 11th
  International Joint Conference on Natural Language Processing (Volume 1: Long
  Papers)}, 2021, pp. 7088--7105.

\bibitem{Boudjellal2021ABioNERAB}
N.~Boudjellal, H.~Zhang, A.~Khan, A.~Ahmad, R.~Naseem, J.~Shang, and L.~Dai,
  ``Abioner: A bert-based model for arabic biomedical named-entity
  recognition,'' \emph{Complex.}, vol. 2021, pp. 6\,633\,213:1--6\,633\,213:6,
  2021.

\bibitem{shaalan2014hybrid}
K.~Shaalan and M.~Oudah, ``A hybrid approach to arabic named entity
  recognition,'' \emph{Journal of Information Science}, vol.~40, no.~1, pp.
  67--87, 2014.

\bibitem{dahan2015first}
F.~Dahan, A.~Touir, and H.~Mathkour, ``First order hidden markov model for
  automatic arabic name entity recognition,'' \emph{International Journal of
  Computer Applications}, vol. 123, no.~7, 2015.

\bibitem{mesmia2018casaner}
F.~B. Mesmia, K.~Haddar, N.~Friburger, and D.~Maurel, ``Casaner: Arabic named
  entity recognition tool,'' in \emph{Intelligent natural language processing:
  Trends and applications}.\hskip 1em plus 0.5em minus 0.4em\relax Springer,
  2018, pp. 173--198.

\bibitem{khalifa2019character}
M.~Khalifa and K.~Shaalan, ``Character convolutions for arabic named entity
  recognition with long short-term memory networks,'' \emph{Computer Speech \&
  Language}, vol.~58, pp. 335--346, 2019.

\bibitem{elsherif2019arabic}
H.~M. Elsherif, K.~M. Alomari, A.~Q.~M. AlHamad, and K.~Shaalan, ``Arabic
  rule-based named entity recognition system using gate.'' in \emph{MLDM (1)},
  2019, pp. 1--15.

\bibitem{alsaaran2021arabic}
N.~Alsaaran and M.~Alrabiah, ``Arabic named entity recognition: A bert-bgru
  approach,'' \emph{Comput. Mater. Contin}, vol.~68, pp. 471--485, 2021.

\bibitem{rom2021supporting}
A.~Rom and K.~Bar, ``Supporting undotted arabic with pre-trained language
  models,'' in \emph{Proceedings of The Fourth International Conference on
  Natural Language and Speech Processing (ICNLSP 2021)}, 2021, pp. 89--94.

\bibitem{ruder2017overview}
S.~Ruder, ``An overview of multi-task learning in deep neural networks,''
  \emph{arXiv e-prints}, pp. arXiv--1706, 2017.

\bibitem{zhang2018overview}
Y.~Zhang and Q.~Yang, ``An overview of multi-task learning,'' \emph{National
  Science Review}, vol.~5, no.~1, pp. 30--43, 2018.

\bibitem{worsham2020multi}
J.~Worsham and J.~Kalita, ``Multi-task learning for natural language processing
  in the 2020s: where are we going?'' \emph{Pattern Recognition Letters}, vol.
  136, pp. 120--126, 2020.

\bibitem{jarrar2022wojood}
M.~Jarrar, M.~Khalilia, and S.~Ghanem, ``Wojood: Nested arabic named entity
  corpus and recognition using bert,'' in \emph{Proceedings of the
  International Conference on Language Resources and Evaluation (LREC 2022),
  Marseille, France, June}, 2022.

\bibitem{ahmed2022tafsir}
S.~Ahmed, R.~van~der Goot, M.~Rehman, C.~Kruse, {\"O}.~{\"O}zsoy, A.~Mehler,
  and G.~Roig, ``Tafsir dataset: A novel multi-task benchmark for named entity
  recognition and topic modeling in classical arabic literature,'' in
  \emph{Proceedings of the 29th International Conference on Computational
  Linguistics}, 2022, pp. 3753--3768.

\bibitem{houlsby2019parameter}
N.~Houlsby, A.~Giurgiu, S.~Jastrzebski, B.~Morrone, Q.~De~Laroussilhe,
  A.~Gesmundo, M.~Attariyan, and S.~Gelly, ``Parameter-efficient transfer
  learning for nlp,'' in \emph{International Conference on Machine
  Learning}.\hskip 1em plus 0.5em minus 0.4em\relax PMLR, 2019, pp. 2790--2799.

\bibitem{zhuang2020comprehensive}
F.~Zhuang, Z.~Qi, K.~Duan, D.~Xi, Y.~Zhu, H.~Zhu, H.~Xiong, and Q.~He, ``A
  comprehensive survey on transfer learning,'' \emph{Proceedings of the IEEE},
  vol. 109, no.~1, pp. 43--76, 2020.

\bibitem{al2020transfer}
M.~Al-Smadi, S.~Al-Zboon, Y.~Jararweh, and P.~Juola, ``Transfer learning for
  arabic named entity recognition with deep neural networks,'' \emph{Ieee
  Access}, vol.~8, pp. 37\,736--37\,745, 2020.

\bibitem{yang2019multilingual}
Y.~Yang, D.~Cer, A.~Ahmad, M.~Guo, J.~Law, N.~Constant, G.~H. Abrego, S.~Yuan,
  C.~Tar, Y.-H. Sung \emph{et~al.}, ``Multilingual universal sentence encoder
  for semantic retrieval,'' \emph{arXiv preprint arXiv:1907.04307}, 2019.

\bibitem{alotaibi2014hybrid}
F.~Alotaibi and M.~Lee, ``A hybrid approach to features representation for
  fine-grained arabic named entity recognition,'' in \emph{Proceedings of
  COLING 2014, the 25th International Conference on Computational Linguistics:
  Technical Papers}, 2014, pp. 984--995.

\bibitem{lan2020empirical}
W.~Lan, Y.~Chen, W.~Xu, and A.~Ritter, ``An empirical study of pre-trained
  transformers for arabic information extraction,'' \emph{arXiv preprint
  arXiv:2004.14519}, 2020.

\bibitem{li2020unified}
X.~Li, J.~Feng, Y.~Meng, Q.~Han, F.~Wu, and J.~Li, ``A unified mrc framework
  for named entity recognition,'' in \emph{Proceedings of the 58th Annual
  Meeting of the Association for Computational Linguistics}, 2020, pp.
  5849--5859.

\bibitem{yu2020named}
J.~Yu, B.~Bohnet, and M.~Poesio, ``Named entity recognition as dependency
  parsing,'' in \emph{Proceedings of the 58th Annual Meeting of the Association
  for Computational Linguistics}, 2020, pp. 6470--6476.

\bibitem{li2022unified}
J.~Li, H.~Fei, J.~Liu, S.~Wu, M.~Zhang, C.~Teng, D.~Ji, and F.~Li, ``Unified
  named entity recognition as word-word relation classification,'' in
  \emph{Proceedings of the AAAI Conference on Artificial Intelligence},
  vol.~36, no.~10, 2022, pp. 10\,965--10\,973.

\bibitem{gu-etal-2022-delving}
\BIBentryALTinterwordspacing
Y.~Gu, X.~Qu, Z.~Wang, Y.~Zheng, B.~Huai, and N.~J. Yuan, ``Delving deep into
  regularity: A simple but effective method for {C}hinese named entity
  recognition,'' in \emph{Findings of the Association for Computational
  Linguistics: NAACL 2022}.\hskip 1em plus 0.5em minus 0.4em\relax Seattle,
  United States: Association for Computational Linguistics, Jul. 2022, pp.
  1863--1873. [Online]. Available:
  \url{https://aclanthology.org/2022.findings-naacl.143}
\BIBentrySTDinterwordspacing

\bibitem{shen2021locate}
Y.~Shen, X.~Ma, Z.~Tan, S.~Zhang, W.~Wang, and W.~Lu, ``Locate and label: A
  two-stage identifier for nested named entity recognition,'' in
  \emph{Proceedings of the 59th Annual Meeting of the Association for
  Computational Linguistics and the 11th International Joint Conference on
  Natural Language Processing (Volume 1: Long Papers)}, 2021, pp. 2782--2794.

\bibitem{xia2019multi}
C.~Xia, C.~Zhang, T.~Yang, Y.~Li, N.~Du, X.~Wu, W.~Fan, F.~Ma, and S.~Y.
  Philip, ``Multi-grained named entity recognition,'' in \emph{Proceedings of
  the 57th Annual Meeting of the Association for Computational Linguistics},
  2019, pp. 1430--1440.

\bibitem{lu2022unified}
Y.~Lu, Q.~Liu, D.~Dai, X.~Xiao, H.~Lin, X.~Han, L.~Sun, and H.~Wu, ``Unified
  structure generation for universal information extraction,'' in
  \emph{Proceedings of the 60th Annual Meeting of the Association for
  Computational Linguistics (Volume 1: Long Papers)}, 2022, pp. 5755--5772.

\bibitem{raffel2020exploring}
C.~Raffel, N.~Shazeer, A.~Roberts, K.~Lee, S.~Narang, M.~Matena, Y.~Zhou,
  W.~Li, P.~J. Liu \emph{et~al.}, ``Exploring the limits of transfer learning
  with a unified text-to-text transformer.'' \emph{J. Mach. Learn. Res.},
  vol.~21, no. 140, pp. 1--67, 2020.

\bibitem{lewis2020bart}
M.~Lewis, Y.~Liu, N.~Goyal, M.~Ghazvininejad, A.~Mohamed, O.~Levy, V.~Stoyanov,
  and L.~Zettlemoyer, ``Bart: Denoising sequence-to-sequence pre-training for
  natural language generation, translation, and comprehension,'' in
  \emph{Proceedings of the 58th Annual Meeting of the Association for
  Computational Linguistics}, 2020, pp. 7871--7880.

\bibitem{yan2021unified}
H.~Yan, T.~Gui, J.~Dai, Q.~Guo, Z.~Zhang, and X.~Qiu, ``A unified generative
  framework for various ner subtasks,'' in \emph{Proceedings of the 59th Annual
  Meeting of the Association for Computational Linguistics and the 11th
  International Joint Conference on Natural Language Processing (Volume 1: Long
  Papers)}, 2021, pp. 5808--5822.

\bibitem{chen2021lightner}
X.~Chen, N.~Zhang, L.~Li, X.~Xie, S.~Deng, C.~Tan, F.~Huang, L.~Si, and
  H.~Chen, ``Lightner: A lightweight generative framework with prompt-guided
  attention for low-resource ner,'' \emph{arXiv preprint arXiv:2109.00720},
  2021.

\bibitem{goo2018slot}
C.-W. Goo, G.~Gao, Y.-K. Hsu, C.-L. Huo, T.-C. Chen, K.-W. Hsu, and Y.-N. Chen,
  ``Slot-gated modeling for joint slot filling and intent prediction,'' in
  \emph{Proceedings of the 2018 Conference of the North American Chapter of the
  Association for Computational Linguistics: Human Language Technologies,
  Volume 2 (Short Papers)}, 2018, pp. 753--757.

\bibitem{chen2019bert}
Q.~Chen, Z.~Zhuo, and W.~Wang, ``Bert for joint intent classification and slot
  filling,'' \emph{arXiv preprint arXiv:1902.10909}, 2019.

\bibitem{weld2021survey}
H.~Weld, X.~Huang, S.~Long, J.~Poon, and S.~C. Han, ``A survey of joint intent
  detection and slot filling models in natural language understanding,''
  \emph{ACM Computing Surveys (CSUR)}, 2021.

\bibitem{mai2018empirical}
K.~Mai, T.-H. Pham, M.~T. Nguyen, T.~D. Nguyen, D.~Bollegala, R.~Sasano, and
  S.~Sekine, ``An empirical study on fine-grained named entity recognition,''
  in \emph{Proceedings of the 27th International Conference on Computational
  Linguistics}, 2018, pp. 711--722.

\bibitem{peng2020toward}
M.~Peng, R.~Ma, Q.~Zhang, L.~Zhao, M.~Wei, C.~Sun, and X.-J. Huang, ``Toward
  recognizing more entity types in ner: an efficient implementation using only
  entity lexicons,'' in \emph{Findings of the Association for Computational
  Linguistics: EMNLP 2020}, 2020, pp. 678--688.

\bibitem{wan2022nested}
J.~Wan, D.~Ru, W.~Zhang, and Y.~Yu, ``Nested named entity recognition with
  span-level graphs,'' in \emph{Proceedings of the 60th Annual Meeting of the
  Association for Computational Linguistics (Volume 1: Long Papers)}, 2022, pp.
  892--903.

\bibitem{lou2022nested}
C.~Lou, S.~Yang, and K.~Tu, ``Nested named entity recognition as latent
  lexicalized constituency parsing,'' in \emph{Proceedings of the 60th Annual
  Meeting of the Association for Computational Linguistics (Volume 1: Long
  Papers)}, 2022, pp. 6183--6198.

\bibitem{li2021span}
F.~Li, Z.~Lin, M.~Zhang, and D.~Ji, ``A span-based model for joint overlapped
  and discontinuous named entity recognition,'' in \emph{Proceedings of the
  59th Annual Meeting of the Association for Computational Linguistics and the
  11th International Joint Conference on Natural Language Processing (Volume 1:
  Long Papers)}, 2021, pp. 4814--4828.

\bibitem{wang2021discontinuous}
Y.~Wang, B.~Yu, H.~Zhu, T.~Liu, N.~Yu, and L.~Sun, ``Discontinuous named entity
  recognition as maximal clique discovery,'' in \emph{Proceedings of the 59th
  Annual Meeting of the Association for Computational Linguistics and the 11th
  International Joint Conference on Natural Language Processing (Volume 1: Long
  Papers)}, 2021, pp. 764--774.

\bibitem{dai2020effective}
X.~Dai, S.~Karimi, B.~Hachey, and C.~Paris, ``An effective transition-based
  model for discontinuous ner,'' in \emph{Proceedings of the 58th Annual
  Meeting of the Association for Computational Linguistics}, 2020, pp.
  5860--5870.

\bibitem{liu2022toe}
J.~Liu, D.~Ji, J.~Li, D.~Xie, C.~Teng, L.~Zhao, and F.~Li, ``Toe: A
  grid-tagging discontinuous ner model enhanced by embedding tag/word relations
  and more fine-grained tags,'' \emph{IEEE/ACM Transactions on Audio, Speech,
  and Language Processing}, vol.~31, pp. 177--187, 2022.

\bibitem{dai2020analysis}
X.~Dai and H.~Adel, ``An analysis of simple data augmentation for named entity
  recognition,'' in \emph{Proceedings of the 28th International Conference on
  Computational Linguistics}, 2020, pp. 3861--3867.

\bibitem{ding2020daga}
B.~Ding, L.~Liu, L.~Bing, C.~Kruengkrai, T.~H. Nguyen, S.~Joty, L.~Si, and
  C.~Miao, ``Daga: Data augmentation with a generation approach for
  low-resource tagging tasks,'' in \emph{Proceedings of the 2020 Conference on
  Empirical Methods in Natural Language Processing (EMNLP)}, 2020, pp.
  6045--6057.

\bibitem{liu22low}
J.~Liu, Y.~Chen, and J.~Xu, ``Low-resource ner by data augmentation with
  prompting,'' in \emph{Proceedings of the Thirty-First International Joint
  Conference on Artificial Intelligence, IJCAI-22}, pp. 4252--4258.

\bibitem{cui2021template}
L.~Cui, Y.~Wu, J.~Liu, S.~Yang, and Y.~Zhang, ``Template-based named entity
  recognition using bart,'' in \emph{Findings of the Association for
  Computational Linguistics: ACL-IJCNLP 2021}, 2021, pp. 1835--1845.

\bibitem{lee2022good}
D.-H. Lee, A.~Kadakia, K.~Tan, M.~Agarwal, X.~Feng, T.~Shibuya, R.~Mitani,
  T.~Sekiya, J.~Pujara, and X.~Ren, ``Good examples make a faster learner:
  Simple demonstration-based learning for low-resource ner,'' in
  \emph{Proceedings of the 60th Annual Meeting of the Association for
  Computational Linguistics (Volume 1: Long Papers)}, 2022, pp. 2687--2700.

\bibitem{yang2018distantly}
Y.~Yang, W.~Chen, Z.~Li, Z.~He, and M.~Zhang, ``Distantly supervised ner with
  partial annotation learning and reinforcement learning,'' in
  \emph{Proceedings of the 27th International Conference on Computational
  Linguistics}, 2018, pp. 2159--2169.

\bibitem{peng2019distantly}
M.~Peng, X.~Xing, Q.~Zhang, J.~Fu, and X.-J. Huang, ``Distantly supervised
  named entity recognition using positive-unlabeled learning,'' in
  \emph{Proceedings of the 57th Annual Meeting of the Association for
  Computational Linguistics}, 2019, pp. 2409--2419.

\bibitem{liang2020bond}
C.~Liang, Y.~Yu, H.~Jiang, S.~Er, R.~Wang, T.~Zhao, and C.~Zhang, ``Bond:
  Bert-assisted open-domain named entity recognition with distant
  supervision,'' in \emph{Proceedings of the 26th ACM SIGKDD International
  Conference on Knowledge Discovery \& Data Mining}, 2020, pp. 1054--1064.

\bibitem{meng2021distantly}
Y.~Meng, Y.~Zhang, J.~Huang, X.~Wang, Y.~Zhang, H.~Ji, and J.~Han,
  ``Distantly-supervised named entity recognition with noise-robust learning
  and language model augmented self-training,'' in \emph{Proceedings of the
  2021 Conference on Empirical Methods in Natural Language Processing}, 2021,
  pp. 10\,367--10\,378.

\bibitem{zhang2021improving}
X.~Zhang, B.~Yu, T.~Liu, Z.~Zhang, J.~Sheng, X.~Mengge, and H.~Xu, ``Improving
  distantly-supervised named entity recognition with self-collaborative
  denoising learning,'' in \emph{Proceedings of the 2021 Conference on
  Empirical Methods in Natural Language Processing}, 2021, pp.
  10\,746--10\,757.

\bibitem{qu2022distantly}
X.~Qu, J.~Zeng, D.~Liu, Z.~Wang, B.~Huai, and P.~Zhou, ``Distantly-supervised
  named entity recognition with adaptive teacher learning and fine-grained
  student ensemble,'' \emph{arXiv preprint arXiv:2212.06522}, 2022.

\bibitem{jiang2021named}
H.~Jiang, D.~Zhang, T.~Cao, B.~Yin, and T.~Zhao, ``Named entity recognition
  with small strongly labeled and large weakly labeled data,'' in
  \emph{Proceedings of the 59th Annual Meeting of the Association for
  Computational Linguistics and the 11th International Joint Conference on
  Natural Language Processing (Volume 1: Long Papers)}, 2021, pp. 1775--1789.

\bibitem{jain2019entity}
A.~Jain, B.~Paranjape, and Z.~C. Lipton, ``Entity projection via machine
  translation for cross-lingual ner,'' in \emph{Proceedings of the 2019
  Conference on Empirical Methods in Natural Language Processing and the 9th
  International Joint Conference on Natural Language Processing
  (EMNLP-IJCNLP)}, 2019, pp. 1083--1092.

\bibitem{feng2021survey}
S.~Y. Feng, V.~Gangal, J.~Wei, S.~Chandar, S.~Vosoughi, T.~Mitamura, and
  E.~Hovy, ``A survey of data augmentation approaches for nlp,'' in
  \emph{Findings of the Association for Computational Linguistics: ACL-IJCNLP
  2021}, 2021, pp. 968--988.

\bibitem{keung2019adversarial}
P.~Keung, Y.~Lu, and V.~Bhardwaj, ``Adversarial learning with contextual
  embeddings for zero-resource cross-lingual classification and ner,'' in
  \emph{Proceedings of the 2019 Conference on Empirical Methods in Natural
  Language Processing and the 9th International Joint Conference on Natural
  Language Processing (EMNLP-IJCNLP)}, 2019, pp. 1355--1360.

\bibitem{chen2021advpicker}
W.~Chen, H.~Jiang, Q.~Wu, B.~Karlsson, and Y.~Guan, ``Advpicker: Effectively
  leveraging unlabeled data via adversarial discriminator for cross-lingual
  ner,'' in \emph{Proceedings of the 59th Annual Meeting of the Association for
  Computational Linguistics and the 11th International Joint Conference on
  Natural Language Processing (Volume 1: Long Papers)}, 2021, pp. 743--753.

\bibitem{lample2019cross}
G.~Lample and A.~Conneau, ``Cross-lingual language model pretraining,''
  \emph{arXiv preprint arXiv:1901.07291}, 2019.

\bibitem{xue2021mt5}
L.~Xue, N.~Constant, A.~Roberts, M.~Kale, R.~Al-Rfou, A.~Siddhant, A.~Barua,
  and C.~Raffel, ``mt5: A massively multilingual pre-trained text-to-text
  transformer,'' in \emph{Proceedings of the 2021 Conference of the North
  American Chapter of the Association for Computational Linguistics: Human
  Language Technologies}, 2021, pp. 483--498.

\bibitem{ebrahimi2021adapt}
A.~Ebrahimi and K.~Kann, ``How to adapt your pretrained multilingual model to
  1600 languages,'' in \emph{Proceedings of the 59th Annual Meeting of the
  Association for Computational Linguistics and the 11th International Joint
  Conference on Natural Language Processing (Volume 1: Long Papers)}, 2021, pp.
  4555--4567.

\bibitem{pires2019multilingual}
T.~Pires, E.~Schlinger, and D.~Garrette, ``How multilingual is multilingual
  bert?'' in \emph{Proceedings of the 57th Annual Meeting of the Association
  for Computational Linguistics}, 2019, pp. 4996--5001.

\bibitem{sil2013re}
A.~Sil and A.~Yates, ``Re-ranking for joint named-entity recognition and
  linking,'' in \emph{Proceedings of the 22nd ACM international conference on
  Information \& Knowledge Management}, 2013, pp. 2369--2374.

\bibitem{luo2015joint}
G.~Luo, X.~Huang, C.-Y. Lin, and Z.~Nie, ``Joint entity recognition and
  disambiguation,'' in \emph{Proceedings of the 2015 Conference on Empirical
  Methods in Natural Language Processing}, 2015, pp. 879--888.

\bibitem{martins2019joint}
P.~H. Martins, Z.~Marinho, and A.~F. Martins, ``Joint learning of named entity
  recognition and entity linking,'' \emph{ACL 2019}, p. 190, 2019.

\bibitem{zelenko2003kernel}
D.~Zelenko, C.~Aone, and A.~Richardella, ``Kernel methods for relation
  extraction,'' \emph{Journal of machine learning research}, vol.~3, no. Feb,
  pp. 1083--1106, 2003.

\bibitem{chan2011exploiting}
Y.~S. Chan and D.~Roth, ``Exploiting syntactico-semantic structures for
  relation extraction,'' in \emph{Proceedings of the 49th Annual Meeting of the
  Association for Computational Linguistics: Human Language Technologies},
  2011, pp. 551--560.

\bibitem{miwa2016end}
M.~Miwa and M.~Bansal, ``End-to-end relation extraction using lstms on
  sequences and tree structures,'' in \emph{Proceedings of the 54th Annual
  Meeting of the Association for Computational Linguistics (Volume 1: Long
  Papers)}, 2016, pp. 1105--1116.

\bibitem{wang2020two}
J.~Wang and W.~Lu, ``Two are better than one: Joint entity and relation
  extraction with table-sequence encoders,'' in \emph{Proceedings of the 2020
  Conference on Empirical Methods in Natural Language Processing (EMNLP)},
  2020, pp. 1706--1721.

\bibitem{abdaoui2021dziribert}
A.~Abdaoui, M.~Berrimi, M.~Oussalah, and A.~Moussaoui, ``Dziribert: a
  pre-trained language model for the algerian dialect,'' \emph{arXiv preprint
  arXiv:2109.12346}, 2021.

\bibitem{el2022adasl}
A.~El~Mekki, A.~El~Mahdaouy, I.~Berrada, and A.~Khoumsi, ``Adasl: An
  unsupervised domain adaptation framework for arabic multi-dialectal sequence
  labeling,'' \emph{Information Processing \& Management}, vol.~59, no.~4, p.
  102964, 2022.

\end{thebibliography}
\end{document}